\documentclass[conference]{IEEEtran}
\IEEEoverridecommandlockouts
\usepackage{cite}
\usepackage{amsmath,amssymb,amsfonts}
\usepackage{algorithmic}
\usepackage{graphicx}
\usepackage{textcomp}
\usepackage{xcolor}
\usepackage{makecell}
\usepackage{booktabs}
\usepackage{multirow}

\usepackage{multirow} 
\usepackage{arydshln}

\usepackage[ruled]{algorithm2e} 

\def\BibTeX{{\rm B\kern-.05em{\sc i\kern-.025em b}\kern-.08em
    T\kern-.1667em\lower.7ex\hbox{E}\kern-.125emX}}
\begin{document}

\title{A Two-Phase Recall-and-Select Framework for Fast Model Selection
}

\author{\IEEEauthorblockN{Jianwei Cui, Wenhang Shi, Honglin Tao, Wei Lu$^{\ast}$, Xiaoyong Du$^{\ast}$ \thanks{*Corresponding author}}
\IEEEauthorblockA{\textit{Renmin University of China, Beijing, China} \\
\{cuijianwei, wenhangshi, honglintao, lu-wei, duyong\}@ruc.edu.cn}
}

\maketitle

\begin{abstract}

As the ubiquity of deep learning in various machine learning applications has amplified, a proliferation of neural network models has been trained and shared on public model repositories. In the context of a targeted machine learning assignment, utilizing an apt source model as a starting point typically outperforms the strategy of training from scratch, particularly with limited training data. Despite the investigation and development of numerous model selection strategies in prior work, the process remains time-consuming, especially given the ever-increasing scale of model repositories. In this paper, we propose a two-phase (coarse-recall and fine-selection) model selection framework, aiming to enhance the efficiency of selecting a robust model by leveraging the models' training performances on benchmark datasets. Specifically, the coarse-recall phase clusters models showcasing similar training performances on benchmark datasets in an offline manner. A light-weight proxy score is subsequently computed between this model cluster and the target dataset, which serves to recall a significantly smaller subset of potential candidate models in a swift manner. In the following fine-selection phase, the final model is chosen by fine-tuning the recalled models on the target dataset with successive halving. To accelerate the process, the final fine-tuning performance of each potential model is predicted by mining the model's convergence trend on the benchmark datasets, which aids in filtering lower performance models more earlier during fine-tuning. Through extensive experimentation on tasks covering natural language processing and computer vision, it has been demonstrated that the proposed methodology facilitates the selection of a high-performing model at a rate about 3x times faster than conventional baseline methods. Our code is available at https://github.com/plasware/two-phase-selection.
\end{abstract}

\begin{IEEEkeywords}
model selection, model clustering
\end{IEEEkeywords}

\section{Introduction}

Nowadays, a plethora of neural networks, meticulously trained in diverse fields such as natural language processing and computer vision, are readily available. These models are commonly hosted on public repositories or model hubs \cite{wolf2019huggingface}\cite{zhao2022tencentpretrain}. Given the wide range of these models' training data, it is plausible that for any specific downstream task, there exists a trained model whose domain distribution of the training dataset is well-transferable for the target task. Employing such a pre-trained model for parameter initialization, followed by fine-tuning on the target dataset often leads to an enhanced performance. This is attributable to the effective transfer and adaptation of the knowledge garnered from the original model to the target task. Therefore, how to select the optimal pre-trained model from a vast collection is crucial for achieving superior training results in a new task, especially when the training data is limited \cite{li2021palette}\cite{renggli2022shift}.




\begin{figure}[t] 

\centerline{\includegraphics[scale = 0.6]{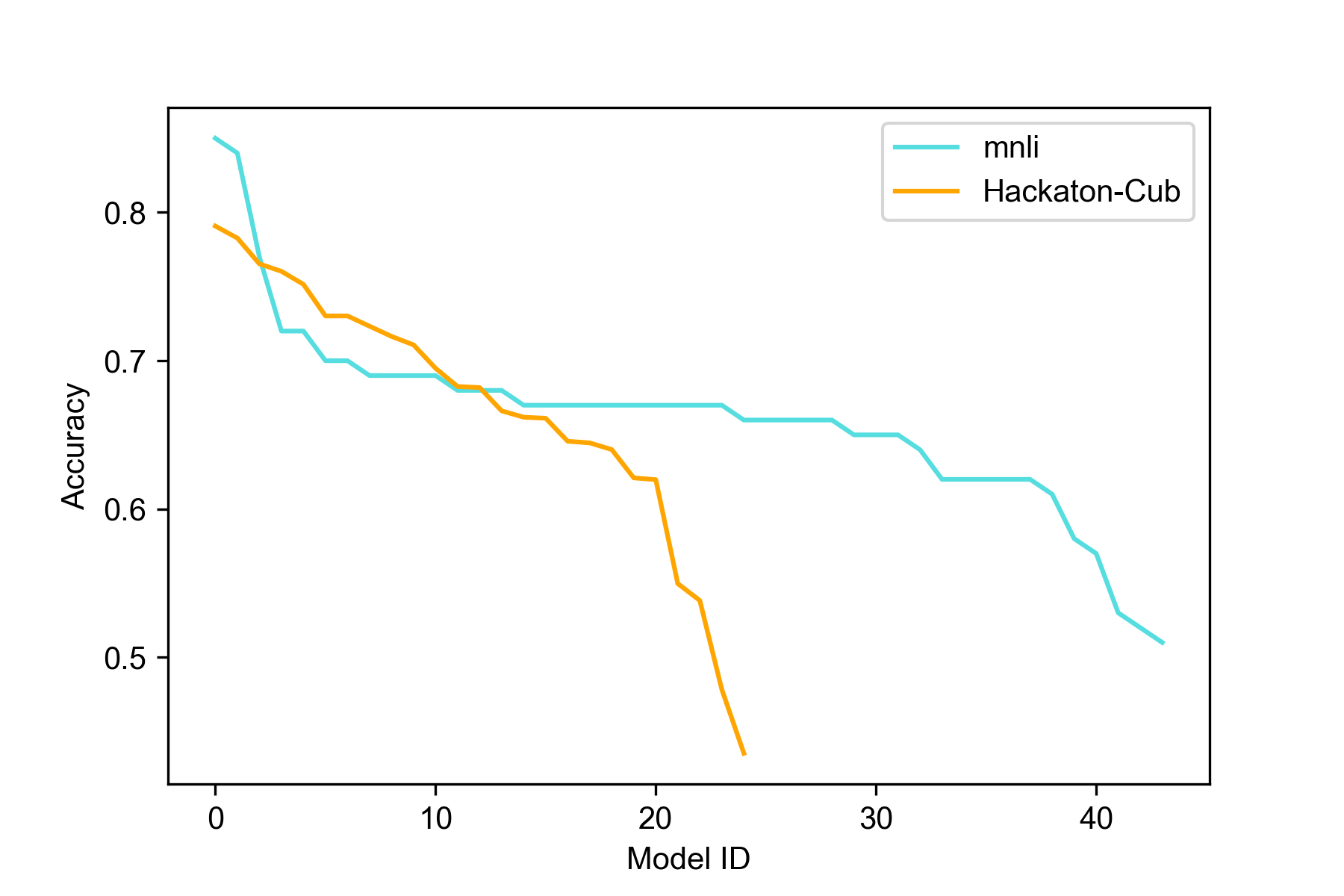}}
\caption{Fine-tuning performance of 44 and 25 pre-trained models on NLP task MNLI \cite{wang2018glue} and CV task CC6204-Hackaton-Cu \cite{WahCUB_200_2011}. The x and y-axis show pre-trained models' ID and their performances on the dataset, respectively. It's noted that for each dataset, the ids are sorted by the model accuracy desc.}
\label{mnli}
\end{figure}

The principal objective of model selection is to efficiently identify a well-suited pre-trained model from a repository for a novel machine learning task. However, the growing repository, though providing potential for improved task initialization, yet escalates the challenge of pinpointing the optimal model. Fig. \ref{mnli} illustrates the fine-tuning results of different models on two distinct machine learning tasks. Although the model pool contains a few models that exhibit commendable performance on the target task, they are markedly outnumbered by models that perform poorly. This discrepancy underscores the increasing complexity of model selection as the volume of available models surges.

The current body of research on this challenge can be bifurcated into two main categories. The first category is centered on the development of lightweight proxy tasks to predict the post-fine-tuning performance of pre-trained models. While the methods offer computational efficiency by obviating the need for direct fine-tuning, they tend to be more prone to selecting sub-optimal models \cite{kornblith2019better}. Conversely, the second category of methods employ a model selection strategy during the fine-tuning process on the target dataset, utilizing a success-halving approach to retain only high-performance models at each training iteration \cite{li2021palette}\cite{renggli2022shift}. However, as illustrated in Fig. \ref{mnli}, only a minor fraction of models in the repository are appropriate for a specific downstream task. Therefore, even with the successive halving strategy, it is computationally inefficient to fine-tune all models in the repository, given that each model needs to be loaded and trained for at least one iteration. Additionally, the efficacy and efficiency of these existing methods decline as the number of pre-trained models continues to expand.

Hence, we propose a two-phase model selection framework, a hybrid approach that amalgamates the advantages of both aforementioned categories, enabling the efficient selection of a suitable pre-trained model for a novel task. We split the model selection to two phases: the first phase, referred as the coarse-recall phase, is designed to identify a handful of promising model candidates based on lightweight proxy tasks. Following this, the second fine-selection phase only necessitates the fine-tuning of models recalled from the first phase to identify the most optimal model. This method significantly improves the efficiency of selecting a suitable model from a large repository, as fine-tuning is only carried out on a substantially reduced subset of models.

The coarse-call phase computes a light proxy task for each model on the target dataset, keeping only the models with high scores for fast filtering. Although the use of proxy tasks avoids fine-tuning, computing a score for each model still makes model loading and inference necessary and inefficient, especially when the model number increases. To speed up, we propose to cluster similar models based on their performances on a set of benchmark datasets. This is inspired by the fact that there is overlap in the training data of public models, so that models that perform similarly on the standard dataset will perform similarly on the new dataset. Specifically, we construct a performance matrix by training each model offline on all benchmark datasets and saving the corresponding performances. Then we cluster the models based on their performance vectors, and each time a new task arrives, we only compute scores for the clusters' representative model. By mining the similarity among models' training, we avoid repeated online computation of proxy scores for similar models and make the model selection more efficient.

\begin{figure*}[t] 
\centerline{\includegraphics[scale = 0.46]{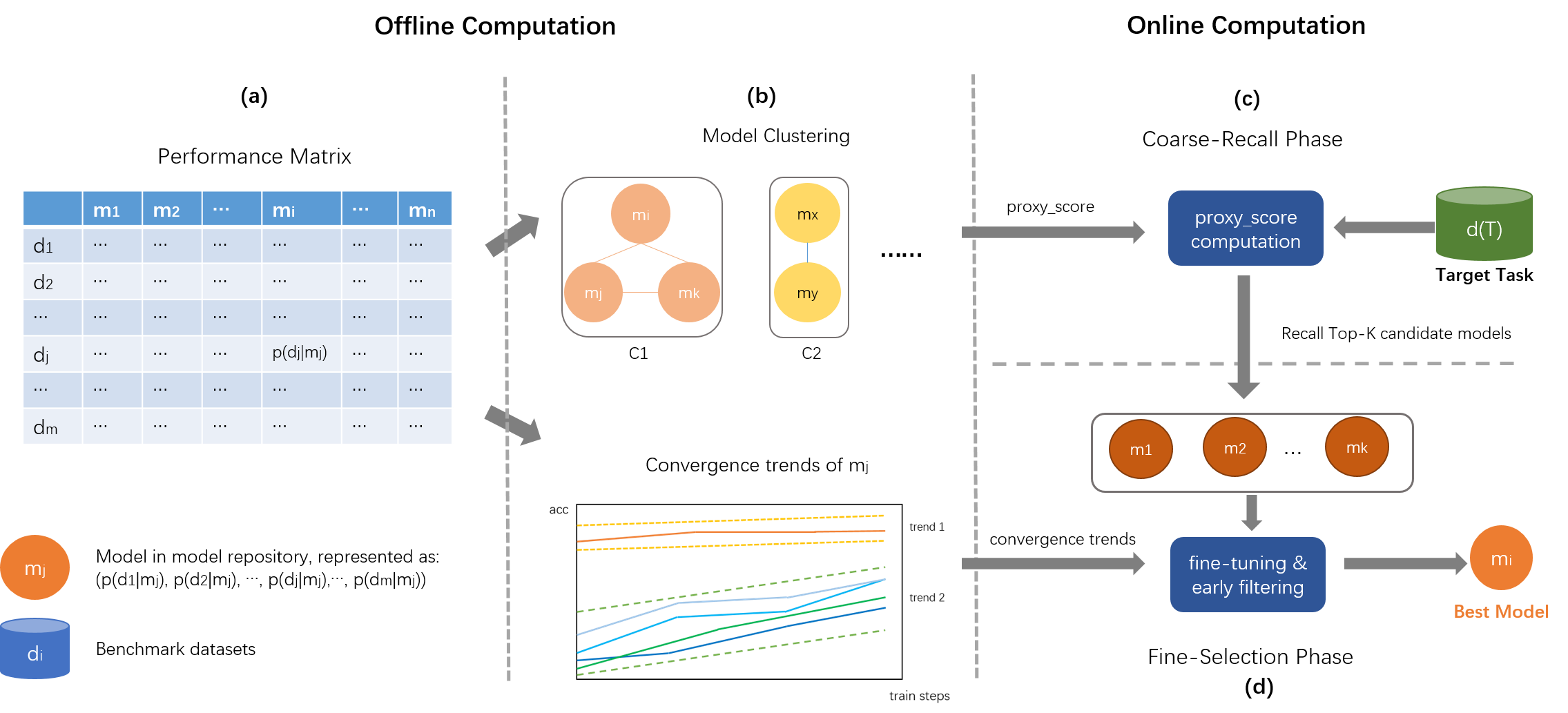}}
\caption{The framework of two-phase model selection: (a) performance matrix, (b) model clustering based on the performance matrix, and convergence trends mining by clustering convergence processes of a pre-trained model on benchmark datasets, (c) coarse-recall phase running recall strategy based on the proxy score computation between a model cluster and the target dataset, and (d) fine-selection phase fine-tunes the recalled models and filters poorly-performance models according to convergence trend. Both (a) and (b) are maintained offline and could be used for any new task.} %
\label{framework}
\end{figure*}

As fine-tuning is more time-consuming, the second fine-selection phase needs to apply more efficient filter strategy to avoid wasting time training low-performance models at early training steps. This is motivated by the model performance consistency at the beginning and end of the training \cite{dodge2020fine}. In this paper, we also apply the successive halving algorithm to filter at least half number of models with lower performance at each training iteration. Meanwhile, as illustrated in Fig. \ref{framework}(b), it is plausible to filter more than half number of models if we can predict the final training performance. Again, we resort to mine the convergence processes between a pre-trained model and benchmark datasets as illustrated in Fig. \ref{framework}(b). Specifically, for every recalled model, we collect the training processes on benchmark datasets, and cluster training processes with similar validation accuracy to form a convergence trend which could predict final test performance range at each iteration step. Then, after the model is fine-tuned on the target dataset for a few steps, we can assign a convergence trend to the model if the current training performance is closed to the training performance of the convergence trend at current step. By this way, the final training performance of a pre-trained model on the target dataset could be predicted more accurate, which could helps to filter more models at early steps.

To this end, we summarize the contributions of this paper as follows:

\begin{itemize}
\item We propose a two-phase model selection framework. The first coarse-recall phase employs a lightweight proxy score to identify a considerably smaller set of promising model candidates. Subsequently, the second fine-selection phase exclusively fine-tunes and filters models from this refined set.


\item To further accelerate the process, we propose mining the training performances of pre-trained models on a collection of benchmark datasets, subsequently clustering similar models. Through clustering, we circumvent computing proxy scores online for each model in the first phase and enhance the accuracy of performance predictions for the left models in the second phase, thereby achieving more efficient and precise selection.

\item We conduct extensive experiments on a substantial variety of training models, encompassing both natural language processing and computer vision domains. The results demonstrate that our proposed framework can effectively identify superior performing models with increasing efficiency, about 3x compared to successive halving and 5x compared to brute force methods.
\end{itemize}

\section{The Framework}
\subsection{Preliminaries}
\textit{Model Repository}. The model repository is a set of pre-trained models, denoted as $M = \{m_1, m_2, ...,m_n\}$. Here, a pre-trained model $m_j$ is a neural network model already trained on an upstream dataset with different learning methods, such as masked language model in natual language processing \cite{devlin2018bert} or image classification for computer vision.

\textit{Benchmark Datasets}. The benchmark datasets comprise representative datasets from the respective domain, such as the GLUE \cite{wang2018glue} for natural language process and various subsets of ImageNet \cite{deng2009imagenet} for computer vision. We use $D = \{d_1, d_2, ..., d_m\}$ to denote the benchmark datasets.

\textit{Performance Matrix}. The performance matrix records the test results of pre-trained models fine-tuned on benchmark datasets, denoted as  $Matrix(D, M)$ with the value $Matrix(D, M)[i][j]$ is the training performance of the pre-trained model $m_j$ on the benchmark dataset $d_i$, also denoted as $p(d_i|m_j)$. The training performance could be measured through different metrics for different tasks, like accuracy for classification tasks.

\textit{Model Cluster}. A model cluster contains a group of models having similar training performances on benchmark datasets. Fig. \ref{framework}(b) illustrates two model clusters, where $C_1$ contains three models ${m_i, m_j, m_k}$ and $C_2$ contains two models ${m_x, m_y}$ respectively.

\textit{Convergence Trend}. The convergence trend clusters datasets into different classes on which the model has a similar training process. We use $CT(m_j)_t$ to denote the class of convergence trend to which the dataset belongs at the $t$ validation for model $m_j$.
Fig. \ref{framework}(b) illustrates two convergence trends for $m_j$. 
The first convergence trend $CT(m_j)_t[0]$ represents a convergence process which could achieve relative higher final training performance, and the second convergence trend $CT(m_j)_t[1]$ achieves lower final training performance. Based on the mined convergence trend, the final performance after training could be predicted more accurate at early training steps.

\textit{Proxy Score}. The proxy score is computed based on the proxy task which predicts the training performance $p(d_i|m_j)$ without actually fine-tuning $m_j$ on $d_i$, denoted as $proxy\_score(d_i|m_j)$. Several proxy tasks have been developed to predict $p(d_i|m_j)$, such as LEEP \cite{nguyen2020leep}, KNN classification \cite{renggli2022model} etc. In this paper, we apply the LEEP score as the $proxy\_score$, which is the average log-likelihood of the expected empirical predictor result of a source model on the target dataset. The advantage of the LEEP score lies in two aspects. Firstly, LEEP could be applied to heterogeneous target tasks that have different label spaces compared with the pre-trained models. Secondly, the computation of the LEEP score does not need extra training, which is more efficient compared with the KNN method.

\textit{Target Task}. We use $T$ to denote the target task and $d(T)$ to denote the training dataset of $T$. The proxy score for model $m_j$ on the target dataset is denoted as $proxy\_score(d(T) | m_j)$, or $proxy\_score(T|m_j)$.

\subsection{Framework Overview}
Fig. \ref{framework} illustrates the whole process of the two-phase framework containing two computation parts:

\textit{Offline}. The construction of the performance matrix constitutes the core offline process, wherein we fine-tune each model in $M$ on the benchmark datasets and record the validation and test results throughout the training process. Subsequently, we execute model clustering based on this matrix, resulting in $MC = \{C_1, C_2, ... , C_p\}$. Although this offline computation is time consuming, this part could be only computed once and then the generated model clusters and intermediate results could be used in the online computation directly.


\textit{Online}. This part implements the two-phase model selection computation for a new task $T$. The coarse-recall phase, denoted as $CR = corase\_recall(M | MC, d(T))$, returns $K$ candidate models from model sets $M$, prone to achieve high training performance on $d(T)$ by computing the proxy score for model clusters $MC$ on $d(T)$. Then, the fine-selection phase, denoted as $fine\_selection(CR |CT, d(T))$, returns the final selected model from the recalled models $CR$ with convergence trend $CT$.

\section{Coarse Recall}
The coarse-recall phase aims to efficiently identify a much smaller number of candidate models which tend to achieve good result on the target task. Fig. \ref{framework}(c) illustrates the overall steps of coarse-recall phase. We firstly present the model clustering process, and then introduce proxy score computation for model clusters on the target dataset to return the recalled models.

\subsection{Model Clustering}
The computation of $proxy\_score(T|m_j)$ needs to load the model into memory and do inference computation on the target dataset. The load and inference step may consume dozens of seconds for a pre-trained model with millions of parameters and a target dataset with hundreds of data items, and consequently is still time consuming. To accelerate the coarse-recall phase, a natural way is to group pre-trained models into clusters by measuring the similarity between pre-trained models, so that the proxy score only needs to be computed for the representative model in a cluster. It reduces the time complexity of coarse-recall phase from $O(|M|)$ to $O(|MC|)$.

The models' similarity measures how two pre-trained models tend to have similar training performance on a target dataset. The training performance could be related to various factors, such as training data domain and quality, model architecture, and parameter size, etc. As these factors are heterogeneous and could not always be available, it maybe unfeasible to combine these factors explicitly to compute the model similarity. In our work, we propose to measure the similarity between models through a data-driven way motivated by the phenomena that models having similar training performances on benchmark datasets also tend to have similar training performance on a new task.

Fig. \ref{framework} (a) and (b) illustrate the model clustering process. Based on the performance matrix $Matrix(D, M)$, each model $m_j$ could be represented as a $\left | D \right | $-dimensional vector $vec(m_{j}) = (p(d_1|m_j), p(d_2|m_j),...,p(d_m|m_j))$. And model clustering could be conducted by any clustering algorithm based on the models' distance $dis(m_{j1}, m_{j2})$ measured based on $vec(m_{j1})$ and $vec(m_{j2})$. As the benchmark datasets cover a group of representative tasks for a machine learning application, the training performances on such datasets could measure both the feature extraction capability and domain characteristics of a pre-trained model. Therefore, for a target task $T$, it is possible that there are benchmark tasks which share similar feature extraction or domain of the training dataset. So the models, which are similar measured by $vec(m_{j1})$ and $vec(m_{j2})$ on benchmark datasets, are also tend to have similar performance in the new task $T$. Here, we measure the model similarity through the average accuracy differences on $k$ benchmark datasets where two models have maximum accuracy differences:
\begin{equation}
sim(m_{j1}, m_{j2}) = 1 - avg(top_k{|vec[m_{j1}] - vec[m_{j2}]|})\label{sim}
\end{equation}

For the performance matrix $Matrix(D, M)$, although there are $m\cdot n$ elements which need $m\cdot n$ times training, it is not necessary to train a pre-trained model on the whole benchmark dataset, since only top accuracy differences will be used to measure the model similarity. Actually, the training performance on a subset of training data with relative small size could be enough. 

Based on the above model similarity measurement, we can adopt state-of-the-art clustering algorithms to group models in model repository, such K-means \cite{kmeans}, hierarchical clustering \cite{hierarchicalclustering}, etc.
After model clustering, for a cluster $C_i$, the model belongs to $C_i$ and has the maximum average training performance on benchmark datasets is selected as the representative model, denoted as $m(C_i)$. Then, the proxy score between the target dataset $d(T)$ and model cluster $C_i$ could be calculated as $proxy\_score(T | m(C_i))$ which could avoid online computing the proxy score for all the models on $d(T)$ in the coarse-recall phase.

\subsection{Model Recall}

Based on the training performance matrix and model clustering result, a $recall\_score$ is computed as Eq. \ref{recall_socre}, where $acc(m_j)$ denotes the average accuracy of $m_j$ on benchmark datasets $D$. We can see the $recall\_score(T|m_j)$ contains two parts. The $acc(m_j)$ counts the prior capacity for a model $m_j$ to any new task, and the $proxy\_score(T|m_j)$ represents how the model $m_j$ matches the specific task $T$. In this paper, we adopt LEEP \cite{nguyen2020leep} to compute the $proxy\_score$ and normalize score between [0, 1]. Combining these two scores, we can sort the models in descendent order, and reserve top $K$ models as the result of the coarse-recall phase.
\begin{equation}
recall\_score(T|m_j) = acc(m_j)\cdot proxy\_score(T | m_j)\label{recall_socre}
\end{equation}

As introduced in previous section, to speed up the computation of coarse-recall phase, we only compute the $proxy\_score$ between the target dataset and the representative model of a cluster, therefore, $proxy\_score(T|m_i)$ could be rewritten as $proxy\_score(T| m(c(m_j))$ where $c(m_j)$ denotes the cluster of model $m_j$ belonging to. Meanwhile, as there may be a number of singleton model clusters ($|Ci| = 1$) after model clustering, the $proxy\_score$ is only computed between the target target $T$ and non-singleton clusters for efficiency consideration. Therefore, for models in non-singleton clusters, the $recall\_score$ is computed as:
\begin{equation}
\begin{split}
\begin{aligned}    
recall&\_score(T | m_j)  = acc(m_j)\cdot \\ 
& proxy\_score(T|m(c(m_j)))\ for\ |c(m_j)| > 1\label{recall_score_2}
\end{aligned}
\end{split}
\end{equation}

As we do not compute the $proxy\_score$ directly for singleton model clusters, the $recall\_score$ for models in singleton clusters is computed as Eq. \ref{recall_score_3} where the $proxy\_score$ is propagated from the representative models of non-singleton clusters (denoted as $C_{non}$) and decayed by the model similarity $sim(m_j, m(C_k))$:
\begin{equation}
\begin{split}
\begin{aligned} 
&recall\_score(T|m_i) = acc(m_i)\cdot \frac{1}{|C_{non}|}\cdot \\ 
&\sum_{k=1}^{|C_{non}|}(sim(m_j, m(C_k))\cdot proxy\_score(T|m(C_k)))\label{recall_score_3}
\end{aligned}
\end{split}
\end{equation}
Combining Eq. \ref{recall_score_2} and Eq. \ref{recall_score_3}, we can compute the $recall\_score$ for all the models in $M$ and return the top $K$ models to the fine-selection phase.


\section{Fine-Selection}
After the initial coarse recall phase, we reduced the number of candidate pre-trained models from $O|M|$ to $O|MC|$. Next, based on successive halving, we utilize the convergence information from fine-tuning the models on benchmark datasets to select a good model more quickly and accurately. 

\subsection{Early Stopping}

\begin{figure}[t] 
\centerline{\includegraphics[scale = 0.39]{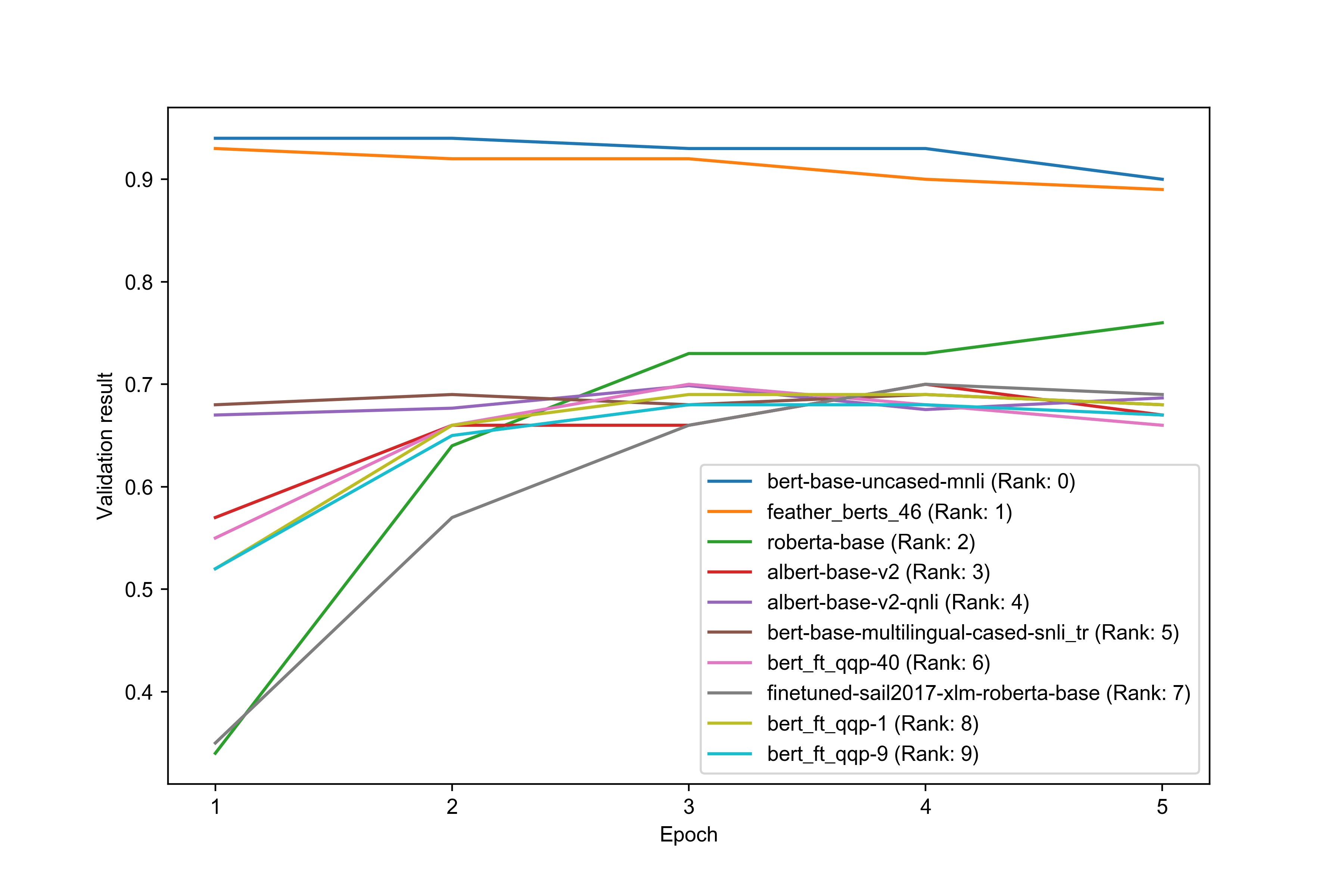}}
\caption{Top-10 models validation and test results on MNLI dataset. Model names ignore the repository name they belong to, the full names can be found in Table \ref{models_nlpcv} in Appendix. B  \cite{two-phase-model-selection-tech-report}.} %
\label{mnli_1}
\end{figure}

Fine-tuning models is a time-consuming process. Key to improve efficiency is the ability to filter out poorly-performing models at an earlier training steps. Therefore, we're interested in understanding whether there is a strong and prevalent correlation between the initial validation performance and the final test performance during the fine-tuning process of pre-trained models. This correlation could potentially allow us to filter out more models at an earlier stage of training. For each target dataset, we plot the validation performance changes during the fine-tuning process for models that pass the initial screening. Fig. \ref{mnli_1} illustrates the performance changes of 10 models screened for the MNLI dataset. It can be observed that the models performing well on the test set also exhibit better validation performance in the early stages of training, and it seems only two models out of the total ten models achieve much higher performance at the first training epoch. Therefore, we do not need to fine-tune all models to the point of convergence. Instead, we can filter out under-performing models at an earlier stage, leading to a more efficient model selection process. Fig. \ref{mnli_2} in Appendix.A \cite{two-phase-model-selection-tech-report} provides  model's performances under another set of hyperparameters, showing the sensitivity of the training process to hyperparameters and the robustness of our method.

\subsection{Successive Halving}
Based on the observation that models which perform well in early training iterations are likely to maintain superior performance when trained to full convergence, successive halving is the state-of-the-art method applied to speed up the model selection process \cite{li2021palette}\cite{renggli2022shift}. The successive halving algorithm operates iteratively in stages, and halves the pool of considered models at each stage. Specifically, each model is trained for a fixed number of iterations at each stage and then its performance is validated. Models with poor performance are discarded, while the better-performing ones proceed to the next round of more intensive training. This process continues until the top models are identified. Assuming that the initial model number is $|M|$ and the training step during each halving is $s$ batches, the total budget for identifying the highest performance model would typically be approximately $|M|\cdot s\cdot log_2(|M|)$ steps. 

\subsection{Fine-Selection Algorithm}

\begin{figure}[t] 
\centerline{\includegraphics[scale = 0.4]{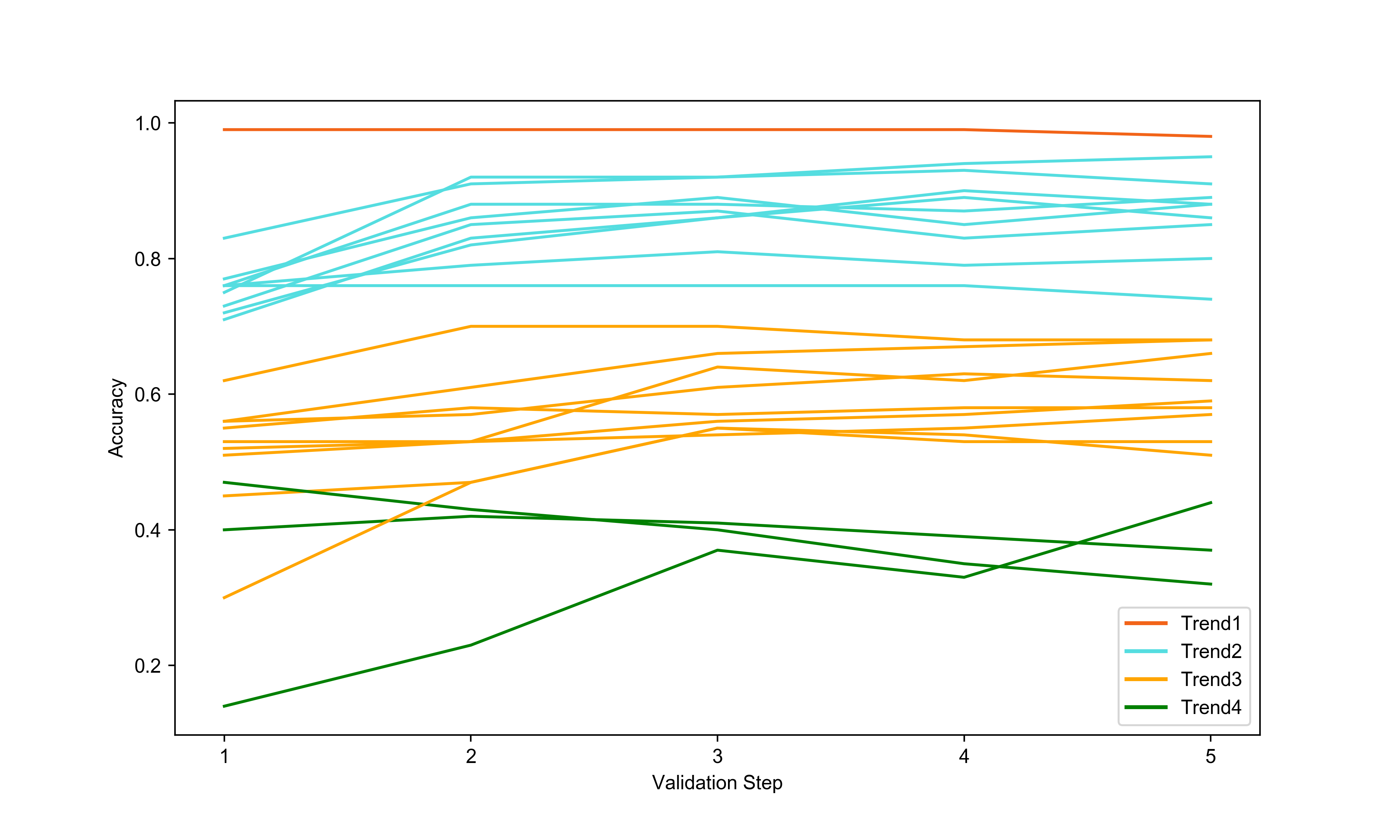}}
\caption{Validation\&Test performance of the DoyyingFace/bert-asian-hate-tweets-asian-unclean-freeze-4 model on 30 datasets.} %
\label{model}
\end{figure}

While successive halving ensures that computational resources are largely devoted to the most promising models, the practice of only filtering out half of the models in each round limits the further improvement of the selection efficiency. Therefore, we propose our refinement method called 'fine-selection', which, built on successive halving, further leverages the fine-tuning information of the model on benchmark datasets. After observing the fine-tuning performance of the model on various datasets, we found that the performance changes of the same model across different datasets can be categorized into distinct clusters. As illustrated in Fig. \ref{model}, the fine-tuning performance of the BERT\_base model on some benchmark datasets can be divided into four groups. Similar phenomena were observed across different models. Therefore, we propose to mining the convergence trend of models from the fine-tuning performance on benchmark datasets to predict the model's final performance on the target dataset.

For a target dataset $d(T)$ and a given pre-trained model $m_j$, after training $m_j$ on $d(T)$ for every $s$ steps, we can compute the validation accuracy $val(T|m_j)_t$ at stage $t$ and predict the final training performance as follows. Firstly, we generate a group of convergence trends for $m_j$, denoted as $\{CT(m_j)_t\}$. Specifically, we cluster the benchmark datasets into $c$ clusters based on the validate accuracy of $m$ on these datasets. Then, a convergence trend $CT(m_j)_t[x] = (\overline{val_x}, \overline{test_x})$ is computed as $\overline{val_x}$ and $\overline{test_x}$ are the average validate and test accuracy of $m_j$ on the datasets in cluster $x$ respectively. Then, based on the generated convergence trends $CT(m_j)_t$, we can assign the best matched convergence trend for $m_j$ trained on $d(T)$ after $s$ steps as Eq. \ref{eq_match}. The final training performance could be predicted as the test accuracy of the matched convergence trend as in Eq. \ref{eq_pred}.

\begin{equation}
\begin{split}
\begin{aligned} 
matched(val(T|m_j)_t) = \underset{x}{\arg\min}({|\overline{val_x} - val(T|m_j)_t|})
\label{eq_match}
\end{aligned}
\end{split}
\end{equation}

\begin{equation}
\begin{split}
\begin{aligned} 
pred(T|m_j)_t = CT(m_j)_t[matched(val(T|m_j)_t)]
\label{eq_pred}
\end{aligned}
\end{split}
\end{equation}

Based on convergence trend mining, Algorithm 1 describes the proposed fine-selection algorithm. Like successive halving, when each remaining model has been fine-tuned for $s$ steps, if the number of remaining models is not 1, fine-selection is performed. The specific process is as follows:
\begin{itemize}
\item Obtaining: Obtain the model's every stage validation and final test results on all benchmark datasets.
\item Convergence Trend Mining and Match: Perform clustering on validation results of current stage, and get the matched convergence trend by Eq. \ref{eq_match}.
\item Predict: Use the mean final test performance of the matched convergence trend as the predicted result by Eq. \ref{eq_pred}.
\item Fine-Filter: Among the remaining models, starting from the model with the worst validation performance, if there exists a model with better validation performance and whose predicted performance is also better by a certain threshold, we remove this model. 
\item Halving: If the number of remaining models is more than half of the number of models at the beginning of the selection, we directly eliminate the model with the worst validation result until the number is reduced by half. 
\end{itemize}

In summary, our fine-selection method ultimately yields a single model fully trained on the target dataset. It ensures that at least half of the models are filtered out at each step, thereby resulting in a selection efficiency significantly higher than that of successive halving.

\begin{algorithm}
\caption{Fine-Selection}
\label{alg:model_selection}
\begin{algorithmic}[1]
\REQUIRE
Recalled models $M_0$; \\
Validation and Test result of $M$ on $D$, $val$ and $test$;\\
Total training steps $T$, validation interval $s$
\ENSURE Final Trained Model 
\FOR{$t = 0$ to $\lfloor T/s \rfloor -1$}
\STATE $M' \gets$ Train each model in $M_t$ 
\IF{$|M'| \neq 1$}
\STATE $v_{j} \gets$ Validate each model
\STATE $x_{j} \gets$ Match convergence trends by Eq. \ref{eq_match}
\STATE $pred_{j} \gets$ Predict final performance by Eq. \ref{eq_pred}
\STATE $M' \gets$ Remove models with lower $v$ and $pred$, \\
 $pred$'s difference should larger than the threshold
\WHILE{$|M'| > \lfloor |M_t|/2 \rfloor$}
\STATE $M' \gets$ Remove models in $M'$ with lowest $v$
\ENDWHILE
\ENDIF
\STATE $M_{t+1} \gets M'$

\ENDFOR
\RETURN $M_{T/s}[0]$
\end{algorithmic}
\end{algorithm}

 

\section{Experiments}
In this section, we study the model selection experiment results on natural language process and computer vision tasks, and demonstrate the effectiveness and efficiency of the proposed methods.

\subsection{Experiment Setup}
\textbf{Pre-trained Models.} The models are collected from HuggingFace \cite{wolf2019huggingface}. For natural language tasks, we select 40 models as model repositories, which both contain the state-of-the-art pre-trained models (such as BERT\cite{devlin2018bert}, Roberta \cite{liu2019roberta}, etc) and the models fine-tuned on some downstream tasks. For computer vision tasks, we select 30 models as repositories. The structure of the models includes ViT \cite{wu2020visual}, BEiT\cite{bao2021beit}, DEiT \cite{touvron2021training}, poolformer \cite{yu2022metaformer}, dinat \cite{hassani2022dilated}, and Visual Attention Network \cite{guo2022visual}. Similar to NLP models, these models contain state-of-the-art models and their fine-tuned versions. A complete list of models are given in Appendix. B \cite{two-phase-model-selection-tech-report}.

\textbf{Datasets.} 
The experiments are conducted on NLP and CV datasets which are divided into benchmark datasets and target datasets. The benchmark datasets are used to construct the performance matrix for model clustering. The target datasets are used to evaluate the proposed two-phase model selection method. The datasets outputs are in the form of classification and are totally different in two parts, which means they are variant in the sub-tasks, label numbers, and label distributions. It is worth noting that our datasets contain both common datasets like GLUE\cite{wang2018glue} and cifar10 \cite{krizhevsky2009learning} as well as domain-specific datasets such as finance and medical science. These datasets are available on HuggingFace and have been split into training and testing sets by their contributors. The datasets used for performance matrix construction and method evaluation are given below:

\begin{itemize}
    \item Benchmark Datasets: The benchmark datasets for NLP are mainly from GLUE\cite{wang2018glue} and SuperGLUE \cite{wang2019superglue}. They are COLA, MRPC, QNLI, QQP, RTE, SST2, STSB and WNLI in GLUE, and CB, COPA, WIC in SuperGLUE. And we use some domain-specific tasks popular in HuggingFace, which are: imdb \cite{maas-EtAl:2011:ACL-HLT2011}, yelp\_review\_full \cite{kim-2014-convolutional}, yahoo\_answer\_topic, dbpedia\_14 \cite{kim-2014-convolutional}, xnli \cite{conneau2018xnli}, anli \cite{nie2019adversarial}, app\_reviews \cite{ZurichOpenRepositoryandArchive:dataset}, trec \cite{li-roth-2002-learning} \cite{hovy-etal-2001-toward}, sick \cite{marelli-etal-2014-sick} and financial\_phrasebank \cite{financial_phrasebank}. For CV benchmark, we use datasets of image classification in different  domains: food101 \cite{food101}, CC6204-Hackaton-Cub-Dataset \cite{WahCUB_200_2011}, cats\_vs\_dogs \cite{cat_vs_dog}, cifar10 \cite{krizhevsky2009learning} and MNIST \cite{lecun1998gradient}.
    \item Target Datasets: For NLP evaluation, four tasks are selected: tweet\_eval \cite{barbieri2020tweeteval} collected from Tweeter reviews, MNLI \cite{williams2017broad} from the GLUE benchmark, MultiRC\cite{multirc}, and Boolq \cite{clark2019boolq} from superGLUE benchmark. For CV evaluation, four image classification datasets with different domains are selected: chest-xray-classification \cite{kermany2018identifying}, MedMNIST \cite{yang2023medmnist}, oxford-flowers \cite{nilsback2008automated}, and beans \cite{beansdata}.
\end{itemize}

We give further description of datasets in Appendix. C\cite{two-phase-model-selection-tech-report}.


\textbf{Performance Matrix.} We build a performance matrix by fine-tuning all the pre-trained models on the benchmark datasets, which contains $40 \times 24$ trains for natural language processing and  $30 \times 10$ trains for computer vision respectively. Each training is conducted with 5 epochs and 4 epochs for natural language processing and computer vision respectively, which is enough to compare the relative accuracy between different trains and could support convergence trend mining for the fine-selection phase. 


\begin{table}[htbp]
\centering
\caption{Clustering Methods Comparison}
\begin{tabular}{ccccc}
\hline
\multirow{2}{*}{Model Similarity} & \multicolumn{2}{c}{Hierarchical clustering} & \multicolumn{2}{c}{K-means} \\
                                  & NLP             & CV             & NLP          & CV           \\ \hline
performance-based                 & \textbf{0.505}  & \textbf{0.806} & 0.466        & 0.702        \\
text-based                   & 0.476           & 0.696          & 0.453        & 0.732        \\ \hline
\end{tabular}
\label{cluster_method_comp}
\end{table}

\subsection{Experiment for Coarse-Recall Phase}

For coarse-recall phase, we first study the experiment results for model clustering and then study the results for model recall.

\textbf{Model Clustering.} We study the model clustering results by comparing different model similarity measurements and clustering algorithms, and the clustering results are measured in terms of silhouette coefficient \cite{rousseeuw1987silhouettes}. For model similarity measurement comparison, the performance-based similarity is calculated by Eq. \ref{sim} and we conduct an experiment in Appendix. D \cite{two-phase-model-selection-tech-report} to determine the parameter $k$. The text-based similarity is calculated from the text of the corresponding model card (Appendix E. \cite{two-phase-model-selection-tech-report} shows an example of a model card). We adopt SBERT\cite{reimers-2019-sentence-bert} to encode the text into an vector so that cosine similarity could be computed. For the clustering algorithms, we compare two state-of-the-art clustering algorithms, K-means and hierarchical clustering.


Table \ref{cluster_method_comp} shows the comparison results. We can see the performance-based similarity achieves higher silhouette coefficient compared to text-based similarity, demonstrating the former could generate a better clustering structure. For clustering algorithm comparison, the hierarchical clustering outperforms the K-means clustering in a clear gap based on performance-based similarity, showing that clusters generated by hierarchical clustering have smaller similarity between and more connection within. To take a step further, it is worth noting that clusters generated by hierarchical clustering are more reasonable than those generated by K-means clustering as discussed below. Therefore, we will conduct the following experiments based on the results of hierarchical clustering.

\begin{table*}[htbp]
\begin{center}
\caption{Model Clustering Results}
\begin{tabular}{|c|c|c|}
\hline
\multicolumn{3}{|c|}{\textbf{Model Clusters of Natural Language Processing}} \\
\hline
\textbf{Cluster} & \textbf{Size}& \textbf{Pre-trained Models} \\
\hline
$C_1$ & 5 & \makecell{Jeevesh8/bert\_ft\_qqp-68, Jeevesh8/bert\_ft\_qqp-9, Jeevesh8/bert\_ft\_qqp-40, \\ connectivity/bert\_ft\_qqp-1, connectivity/bert\_ft\_qqp-7}  \\
\hline
$C_2$ & 7 & \makecell{Jeevesh8/512seq\_len\_6ep\_bert\_ft\_cola-91,  anirudh21/bert-base-uncased-finetuned-qnli, Jeevesh8/bert\_ft\_cola-88, \\ 
manueltonneau/bert-twitter-en-is-hired, bert-base-uncased, \\
aditeyabaral/finetuned-sail2017-xlm-roberta-base, DoyyingFace/bert-asian-hate-tweets-asian-unclean-freeze-4} \\
\hline
$C_3$ & 5 & \makecell{Jeevesh8/feather\_berts\_46, ishan/bert-base-uncased-mnli \\
roberta-base, Alireza1044/albert-base-v2-qnli, albert-base-v2} \\
\hline
$C_4$ & 2 & \makecell{CAMeL-Lab/bert-base-arabic-camelbert-mix-did-nadi, aliosm/sha3bor-metre-detector-arabertv2-base
} \\
\hline
$C_5$ & 2 & \makecell{Splend1dchan/bert-base-uncased-slue-goldtrascription-e3-lr1e-4, aychang/bert-base-cased-trec-coarse
} \\
\hline
$C_6$ & 3 & \makecell{	aviator-neural--bert-base-uncased-sst2, distilbert-base-uncased, \\ 18811449050--bert\_finetuning\_test } \\
\hline
$C_7$ & 4 & \makecell{Jeevesh8/init\_bert\_ft\_qqp-33, Jeevesh8/init\_bert\_ft\_qqp-24, \\ connectivity/bert\_ft\_qqp-17, connectivity/bert\_ft\_qqp-96}  \\
\hline
$C_8$ & 2 & \makecell{XSY/albert-base-v2-imdb-calssification, emrecan/bert-base-multilingual-cased-snli\_tr} \\
\hline
\multicolumn{3}{|c|}{\textbf{Model Clusters of Computer Vision}} \\
\hline
\textbf{Cluster} & \textbf{Size}& \textbf{Pre-trained Models} \\
\hline
$C_1$ & 6 & \makecell{facebook/deit-base-patch16-224, facebook/deit-base-patch16-384, facebook/dino-vits16, \\ facebook/vit-msn-base, facebook/vit-msn-small, Visual-Attention-Network/van-large} \\
\hline
$C_2$ & 2 & \makecell{facebook/deit-small-patch16-224, Visual-Attention-Network/van-base} \\
\hline
$C_3$ & 11 & \makecell{facebook/dino-vitb16, facebook/dino-vitb8, google/vit-base-patch16-224, \\ google/vit-base-patch16-384, lixiqi/beit-base-patch16-224-pt22k-ft22k-finetuned-FER2013-6e-05, \\ lixiqi/beit-base-patch16-224-pt22k-ft22k-finetuned-FER2013-7e-05, \\ lixiqi/beit-base-patch16-224-pt22k-ft22k-finetuned-FER-5e-05-3, \\ microsoft/beit-base-patch16-224, microsoft/beit-base-patch16-224-pt22k-ft22k, microsoft/beit-base-patch16-384, \\ nateraw/vit-age-classifier} \\
\hline
$C_4$ & 2 & \makecell{shi-labs/dinat-large-in22k-in1k-224, shi-labs/dinat-large-in22k-in1k-384} \\
\hline
$C_5$ & 2 & \makecell{sail/poolformer-m36, sail/poolformer-m48} \\
\hline
$C_6$ & 2 & \makecell{shi-labs/dinat-base-in1k-224, microsoft/beit-large-patch16-224-pt22k} \\ 
\hline
\end{tabular}
\label{hierarchical clustering result}
\end{center}
\end{table*}

Table \ref{hierarchical clustering result} illustrates our clustering results of non-singleton clusters for natural language processing and computer vision tasks. For natural language processing models, there are 8 non-singleton clusters which contain 30 models out of a total of 40 models. For computer vision tasks, there are 6 non-singleton models that contain almost all the models. We study the details for some clusters. For natural language process models, as the introduction information is not available for many models, we infer the model training process from the model name. For $C_1$ and $C_2$ clusters, we can see they mainly contain pre-trained models fine-tuned on qqp \cite{sharma2019natural} and cola \cite{warstadt2019neural} datasets respectively, demonstrating the models fine-tuned on the same downstream tasks could be grouped together by model clustering. On the other hand, we can see there are also models with names containing qqp that are not clustered into $C_1$, such as $C_7$, demonstrating the performance of models with similar model names may also vary, which may be caused by different training setups. For $C_3$ cluster, we find that it groups models with names containing mnli and feather\_berts together, and it is reasonable since we can find from \cite{mccoy2019berts} that the feather\_berts models are also BERT models fine-tuned on the MNLI datasets. This also demonstrates the effectiveness of the model clustering. The results of computer vision clusters exhibit a similar phenomenon. The $C_1$ cluster mainly contains base-size deit models \cite{touvron2021training}, small-size vit models using dino \cite{caron2021emerging}, and vit models using msn \cite{assran2022masked}. Looking further into the models, we discover that these three kinds of models are all pre-trained or fine-tuned on dataset Imagenet-1k \cite{russakovsky2015imagenet}. The $C_3$ cluster mainly contains base-size vit models using dino, vit base models, and beit base models. Similar to $C_1$ cluster, these models, except dino-vit, are all pre-trained or fine-tuned with dataset Imagenet-21k \cite{ridnik2021imagenet}. On the other hand, we can also see models that are not grouped in one cluster even though they share similar names or training datasets, like dino vit models in $C_1$ and $C_3$, which also demonstrates models with similar model names may have different performance. We put the result of the K-means clustering method in Appendix. F \cite{two-phase-model-selection-tech-report}. Generally speaking, some clusters contain models of different structures and/or training datasets.

We also study the performance of models in non-singleton clusters in Table \ref{cluster_performance}. We can see the average accuracy of models in non-singleton clusters is significantly higher than that of models in singleton clusters for both natural language and computer vision tasks. Meanwhile, we also compute the count of models that achieve maximum accuracy for a benchmark dataset, and it exhibits a similar result that models in non-singleton clusters almost contribute to all the best models for benchmark datasets. The result of Table \ref{cluster_performance} demonstrates that models in the non-singleton clusters tend to achieve higher training performance. This could be explained by the fact that the high-quality model may achieve similar performance bounded by the state-of-the-art model, while on the other hand, the performance of poorly-performed models may differ a lot on different datasets. This phenomenon supports the $proxy\_score$ computation strategy in the coarse-recall phase that we only compute the $proxy\_score$ between the target dataset and the representative models of non-singleton clusters, and propagate the score to models in singleton clusters.

\begin{table}[htbp]
\begin{center}
\caption{Performance Comparison of models in Singleton and Non-Singleton Clusters}
\begin{tabular}{cccc}
\hline
\textbf{Task Type}   & \textbf{Cluster Type} & \textbf{Avg(Acc)} & \textbf{No. Maximum(Acc)} \\ \hline
\multirow{2}{*}{NLP} & Non-Singleton         & 0.67              & 22                        \\
                     & Singleton             & 0.61              & 2                         \\
\multirow{2}{*}{CV}  & Non-Singleton         & 0.84              & 10                        \\
                     & Singleton             & 0.73              & 0                         \\ \hline
\end{tabular}
\label{cluster_performance}
\end{center}
\end{table}

\begin{figure*}[t] 
\centerline{\includegraphics[scale = 0.45]{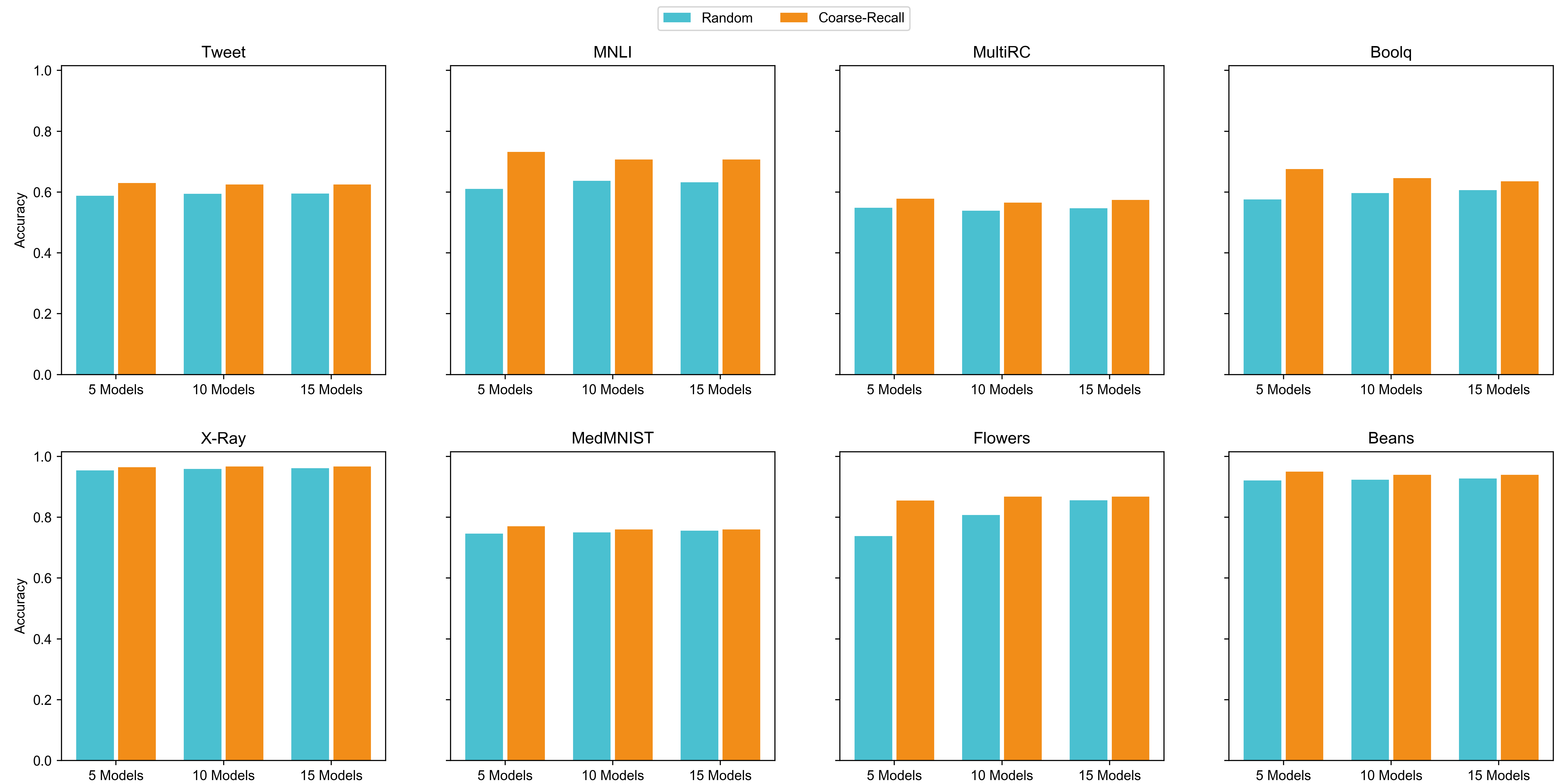}}
\caption{The average accuracy comparison of recalled models} %
\label{avg_acc_recalled}
\end{figure*}

\begin{figure*}[t] 
\centerline{\includegraphics[scale = 0.45]{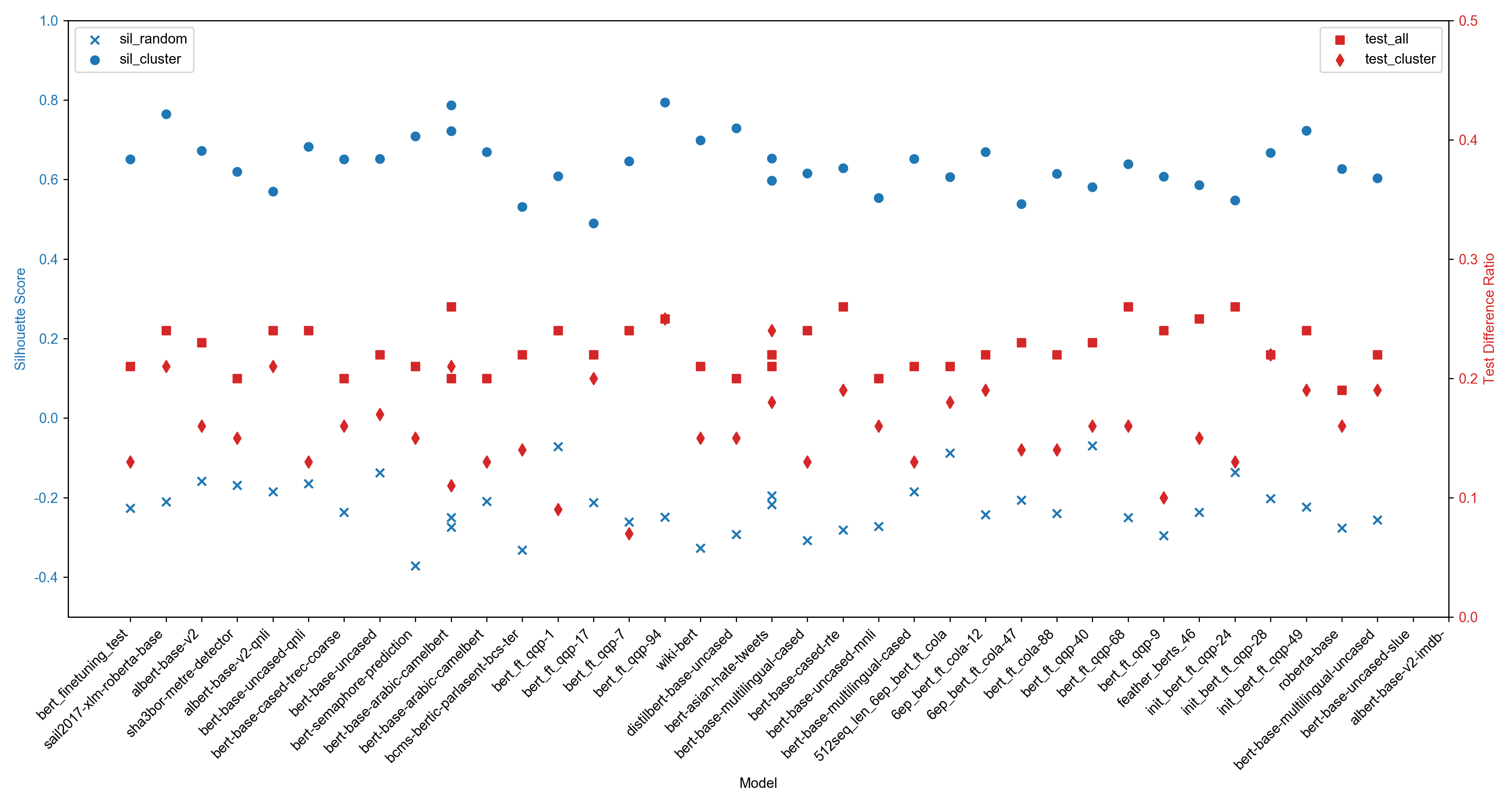}}
\caption{Clustering Performance based on the first validation results. The blue color represents the comparison between random clustering and clustering based on validation performance, with silhouette score as the selected metric. The higher the silhouette score, the better the clustering effect. The red color represents the comparison between the prediction of test performance using the clustering method and the mean of all historical test performances. The metric used here is the absolute difference between the predicted and actual test performance divided by the actual test performance. We calculate the average with each dataset as the target dataset; the smaller the value, the more accurate the prediction performance. Full model name could be found in Table \ref{models_nlpcv} in Appendix.B\cite{two-phase-model-selection-tech-report}.} %
\label{cluster}
\end{figure*}

\textbf{Model Recall.}
To evaluate the effectiveness of coarse-recall phase, we fine-tune all the models on corresponding target datasets to get the actual training performance. Then, we compare the average training accuracy on the target datasets of top $K$ recalled models. Fig. \ref{avg_acc_recalled} compares the average accuracy between coarse-recall and random-recall. Here, random-recall represents randomly return $K$ models from model repository. We can see coarse-recall achieves higher accuracy compared with random-recall on all the eight target datasets. Meanwhile, for coarse-recall, the average accuracy of smaller $K$ values is higher than that of bigger $K$ values, demonstrating the top models recalled by coarse-recall tend to achieve higher training performance compared with models with lower recall score. Meanwhile, we also find that the top 5 models recalled by coarse\_recall has contained the model achieving maximum training performance; and for tweet, beans and MedMNIST datasets, the number of recalled models to contain the best model is 10, 10, 15 respectively. In the following experiments, we empirically set the number of recalled models as 10, which accounts for about 25\% and 30\% of corresponding total models for natural language processing and computer vision tasks respectively.

\subsection{Experiments for Fine-Selection Phase}

\textbf{Convergence Trend.} In this section, we demonstrate the effectiveness of mined convergence trend based on a model's validation performance on benchmark datasets. We apply clustering methods during the fine-selection phase, and usually, a single filtering is sufficient to select the final model. Therefore, we only consider the performance of clustering based on the first validation results. Firstly, we directly measure the effect of clustering through the silhouette coefficient \cite{rousseeuw1987silhouettes}, where a higher value indicates a better clustering outcome. As shown in the blue part of Fig. \ref{cluster}, across all models, clustering based on validation significantly outperforms random clustering, demonstrating the effectiveness of validation results for clustering. In addition, we consider each benchmark dataset as the target dataset, assess the feasibility of predicting the final test performance using the mean test performance of models within the same cluster as the current model belongs to. We compare this with predicting the test performance based on the mean of all benchmark dataset test performances. The final metric is the absolute difference between the predicted and actual test performance divided by the actual test performance, averaged across all datasets. The red part of Fig. \ref{cluster} demonstrates that our method predicts final test performance more accurately. This not only validates the feasibility of using the mean test performance within the cluster as a prediction but also demonstrates the effectiveness of clustering. In summary, it is feasible to select models by mining convergence trend based on validation performance and further predicting the model's final test performance.


\textbf{Filtering Threshold}
Given the validation fluctuations during the model training process and the potential significant discrepancies between benchmark and target datasets, utilizing convergence trends to filter models might eliminate models with good performance. Therefore, we propose introducing a threshold, stipulating that a model is only filtered out when there is another model with better validation and a predicted performance improvement exceeding this threshold. The threshold is a proportion of the difference between the predicted performances. Table \ref{Filter_threshold} illustrates the performance of fine-tuning methods under various threshold conditions. It can be observed that the threshold ensures better-performing models are filtered out later, albeit at the expense of efficiency. In order to more intuitively represent the efficiency of our method and the performance of the selected models, we uniformly use a 0\% threshold in subsequent experiments.

\begin{table}[]
\centering
\caption{Accuracy and Time comparison among different filtering threshold settings in fine-selection. 0\% is the original setting.}
\begin{tabular}{llcccc}
\hline
Models                   &      & 0\%   & 1\%   & 5\%   & 10\%  \\ \hline
\multirow{2}{*}{MNLI}    & Accuracy  & 0.85  & 0.85  & 0.85  & 0.85  \\
                         & RunTime & 14    & 14    & 15    & 16    \\ \hdashline
\multirow{2}{*}{MultiRC} & Accuracy  & 0.63  & 0.63  & 0.63  & 0.63  \\
                         & RunTime & 16    & 16    & 19    & 19    \\ \hdashline
\multirow{2}{*}{Flowers} & Accuracy  & 0.985 & 0.985 & 0.986 & 0.986 \\
                         & RunTime & 15    & 15    & 18    & 18    \\ \hdashline
\multirow{2}{*}{X-Ray}   & Accuracy  & 0.966 & 0.966 & 0.969 & 0.969 \\
                         & RunTime & 14    & 15    & 18    & 18    \\ \hline
\end{tabular}
\label{Filter_threshold}
\end{table}

\textbf{Method Comparison.} After validating the effectiveness of convergence trend mining from the model training performance on benchmark datasets, we further verify its value in model selection methods.

\subsubsection{Performance}
\begin{figure*}[t] 
\centerline{\includegraphics[scale = 0.445]{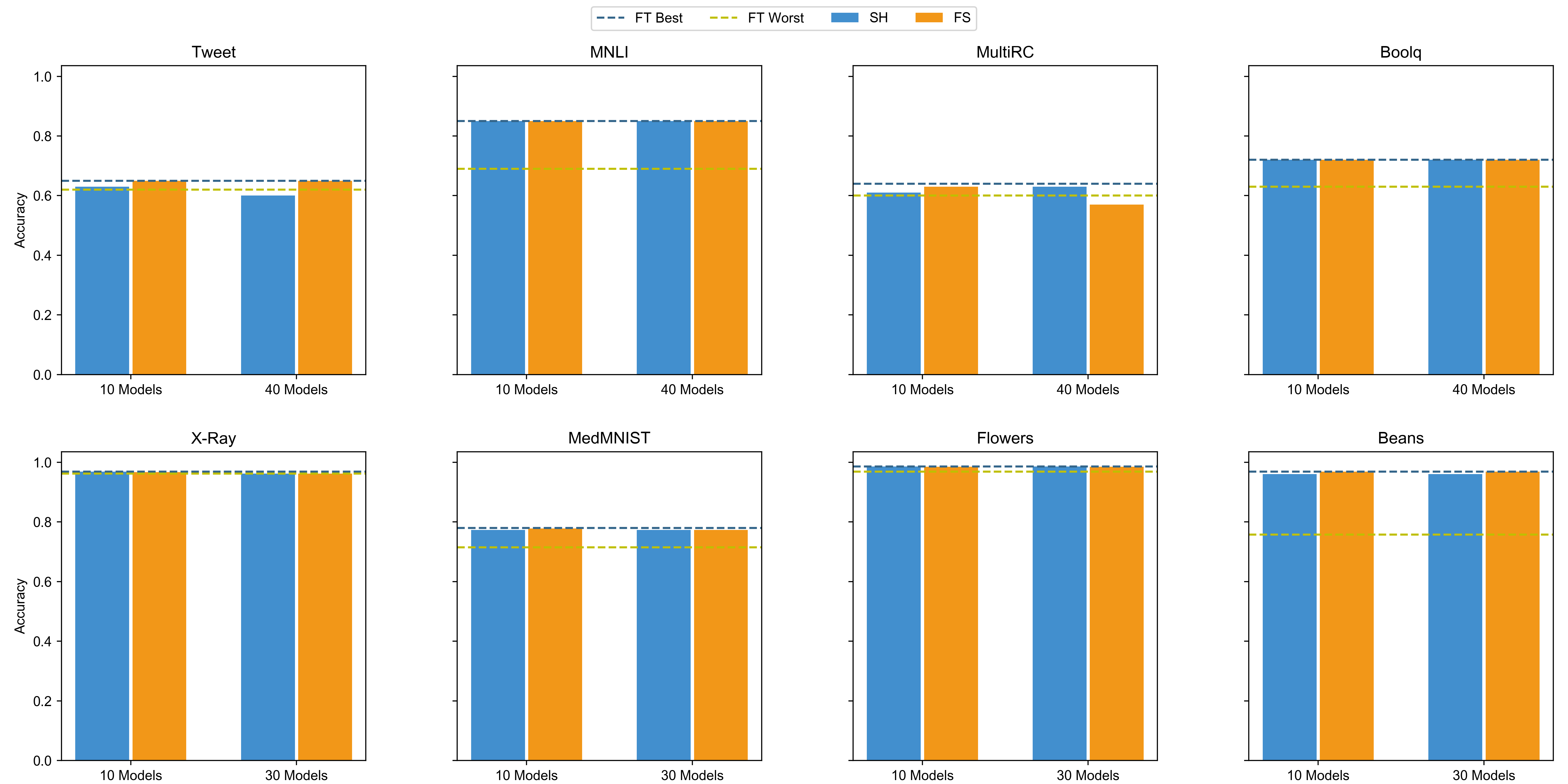}}
\caption{Selected model's performance comparison between successive halving(SH) and our fine-selection(FS) method on NLP datasets: MNLI, Tweet, Boolq, MultiRC and CV dataset: Beans, MedMNIST, Flowers and X-ray. 10, 30 and 40 models represent initial model number for filtering. The best and worst model performances in the top 10 models for each dataset are also provided.} %
\label{performace}
\end{figure*}

Fig. \ref{performace} compares the final performances of the last model selected by the successive halving and our proposed fine-selection method among the top 10 performing and all of the models. At the same time, we provide the best and worst performances among the top 10 models for NLP and CV tasks. We can find that the fine-selection (FS) method is always able to pick out the optimal or near-optimal model. However, the traditional successive halving method may not necessarily select the best model.

 
\subsubsection{Time}
In addition to the performance of the selected model, the speed of model selection is equally important. In Table \ref{time_comp}, we compare the speed improvement of the two methods based on brute-force search (BS), that is, fine-tuning all models for a fixed number of epochs. Given the consistency of training settings and hardware environments, we use the total number of fine-tuning epochs across all models to represent the time for model selection. We can observe that our method has a noticeable efficiency improvement compared to successive halving in all datasets and different model numbers. Furthermore, it's worth noting that the time taken per epoch is considerable, thus the time saved in model selection by our method is significant.

\begin{table}[t]
\begin{center}
\caption{Time comparisons among different model selection methods. NLP tasks are trained for 5 epochs and CV tasks are trained for 4 epochs. Runtime is total training epoch number. Selections of top 10 models and all of the models, whose number is 40 and 30 for NLP and CV, is both provided. }
\begin{tabular}{llcccc}
\hline
Models       & \multicolumn{1}{l}{} & \multicolumn{2}{c}{10}                                                                                                 & \multicolumn{2}{l}{40(NLP)/30(CV)}                                                                                     \\
             & \multicolumn{1}{l}{} & \begin{tabular}[c]{@{}c@{}}Runtime\\ (epoch)\end{tabular} & \begin{tabular}[c]{@{}c@{}}Speedup\\ (vs. BF)\end{tabular} & \begin{tabular}[c]{@{}c@{}}Runtime\\ (epoch)\end{tabular} & \begin{tabular}[c]{@{}c@{}}Speedup\\ (vs. BF)\end{tabular} \\ \hline
\textbf{NLP} & BF                   & 50                                                        &                                                            & 200                                                       &                                                            \\
             & SH                   & 19                                                        & 2.63x                                                      & 77                                                        & 2.60x                                                      \\ \hdashline
Tweet        & FS                   & 14                                                        & 3.57x                                                      & 44                                                        & 4.55x                                                      \\
MNLI         & FS                   & 14                                                        & 3.57x                                                      & 44                                                        & 4.55x                                                      \\
MultiRC      & FS                   & 15                                                        & 3.33x                                                      & 46                                                        & 4.35x                                                      \\
Boolq        & FS                   & 16                                                        & 3.13x                                                      & 48                                                        & 4.17x                                                      \\ \hline
\textbf{CV}  & BF                   & 40                                                        &                                                            & 120                                                       &                                                            \\
             & SH                   & 18                                                        & 2.22x                                                      & 55                                                        & 2.18x                                                      \\ \hdashline
X-Ray        & FS                   & 13                                                        & 3.08x                                                      & 38                                                        & 3.16x                                                      \\
MedMNIST     & FS                   & 15                                                        & 2.67x                                                      & 37                                                        & 3.24x                                                      \\

Flowers       & FS                   & 15                                                        & 2.67x                                                      & 36                                                        & 3.33x      \\
Beans        & FS                   & 17                                                        & 2.35x                                                      & 41                                                        & 2.93x                                                      \\ \hline                                               
\end{tabular}
\label{time_comp}
\end{center}
\end{table}


\subsubsection{Scaling to more models}
In addition to the 10 models obtained based on coarse-recall, we further explored the performance of our method when the number of models increased. In Fig. \ref{performace} and Table \ref{time_comp}, we provided performance and time comparisons for 40 models of natural language process and 30 models of computer vision. We found that as the number of models increases, our method not only maintains the quality of model selection but also significantly saves time. This indicates that the fine-selection method is not only suitable for fine-grained selection after coarse-grained model recall, but it is also effective for a larger number of models. Therefore, our method can be expanded to accommodate more models, making it capable of coping with the increasing emergence of pre-trained models.

\subsection{Overall Performance}

We study the end-to-end model selection effectiveness and efficiency in this section. The comparison methods are also brute-force search (BF) and successive halving (SH) in last section. The number of total models are 40 and 30 for NLP and CV, and recalled model is 10. Table \ref{end2end} shows accuracy and efficiency comparison among different methods where CR+FS stands for the two-phase framework (coarse-recall and fine-selection) in this paper. We can see both SH and two-phase model selection methods proposed in this paper achieve near accuracy compared with BF. Table \ref{end2end} also exhibits the time consumed for the three methods in terms of training epochs. For CR+FS, as $proxy\_score$ needs to be computed in coarse-recall phase which needs to do inference on the target task, we count the computation time as $0.5 \cdot |MC|$ epochs because the inference do not need to compute the gradients to do back-propagation. From Table \ref{end2end}, We can see the two-phase model selection methods achieve about 2x to 3x times faster compared with SH and about 5x to 8x times faster compared with BF. The speed up comes from both phases where the coarse-recall phase largely reduce the number of models need to be fine-tuned and the fine-selection phase further reduce computation time of fine-tuning by exploiting the convergence trend. The results demonstrate the model selection method proposed in this paper could achieve high training performance with significantly lower compute time, hence are more suitable to address large scale model repository. 

\begin{table}[t]
\begin{center}
\caption{End-to-End comparisons on time and performance among different model selection methods. 2PH, BF, SH is short for 2 phase (Coarse Recall +Fine Selection), Brute Forcing, Seccessive Halving, respectively. Acc is accuracy metric.}
\begin{tabular}{lcccccc}
\hline
         & Runtime & \multicolumn{2}{c}{Speedup} & \multicolumn{3}{c}{Acc} \\
         & 2PH   & (vs.BF)     & (vs.SH)     & BF     & SH     & 2PH \\ \hline
Tweet    & 19    & 10.53x        & 4.05x        & 0.650  & 0.60   & 0.650 \\
MNLI     & 19    & 10.53x        & 4.05x        & 0.850  & 0.850  & 0.850 \\
MultiRC  & 20    & 10.00x        & 3.85x        & 0.640  & 0.630  & 0.630 \\
Boolq    & 21    & 9.52x        & 3.67x        & 0.720  & 0.720  & 0.720 \\ \hdashline
X-Ray    & 18    & 6.67x        & 3.06x        & 0.969  & 0.962  & 0.962 \\
MedMNIST & 20    & 6.00x        & 2.75x        & 0.779  & 0.773  & 0.773 \\
Flowers  & 20    & 6.00x        & 2.75x        & 0.986  & 0.986  & 0.985 \\
Beans    & 22    & 5.45x        & 2.50x        & 0.968  & 0.961  & 0.961 \\ \hline
\end{tabular}
\label{end2end}
\end{center}
\end{table}

\begin{table}[t]
\caption{Case study of the final selected model after coarse-recall and fine-selection on four datasets. Acc and R are short for accuracy and rank. The rank for CR are obtained by sorting the recalled models by proxy score. Avg\_Acc is obtained by computing the average accuracies of the recalled models.}
\centering
\begin{tabular}{lcccc}
\hline
Dataset  & Best\_model            & Acc   & R@CR & Avg\_Acc \\ \hline
MultiRC  & albert-base-v2         & \textbf{0.630}  & 5       & 0.574        \\
Boolq    & bert-base-uncased-mnli & \textbf{0.720}  & 0       & 0.635       \\
MedMNIST & vit-base-patch16-384   & \textbf{0.773}  & 1       & 0.768       \\ 
Flowers  & vit-base-patch16-224   & \textbf{0.985}  & 9       & 0.891        \\
\hline
\end{tabular}
\label{case_study}
\end{table}


\subsection{Generalization Study}
The generalization capability of the proposed method for new target tasks lies in three aspects. Firstly, we can address new tasks with different domain distribution and task types compared with benchmark datasets. This is because the benchmark datasets are only used to measure the model similarity offline and will be used neither in corase-recall nor fine-selection phase. Secondly, the LEEP score in the coarse-recall phase proves to be able to measure the transferability between heterogeneous tasks. And thirdly, the fine-selection phase selects model directly based on the fine-tuning results at different validation intervals, which is more accurate to evaluate the final transferability of a source model.

Table \ref{case_study} illustrates the best selected models for four target tasks which have different domain distribution and task type compared with benchmark datasets. The best models are ranked higher at coarse-recall phase and the accuracy are all higher than the average accuracy of all models. The Boolq is a question answering task for yes/no questions based on given passages, and the best model selected for Boolq is bert-base-uncased-mnli which is the bert model fine-tuned on the MNLI dataset. As the input format and output label space are both different for Boolq and MNLI, the result demonstrates the proposed method could capture the latent transferability between heterogeneous tasks. On the other hand, the MultiRC dataset selects albert-base-v2 as the best model, which is a pre-trained model not fine-tuned on any downstream dataset. For computer vision task, both Flowers and MedMNIST datasets select the vit-base-patch16 models which are trained on data with different domain distribution compared with corresponding tasks, demonstrating the out-of-domain capability of proposed method in CV tasks.

\section{Related Work}
Pre-training could be viewed as an application of deep transfer leanring \cite{tan2018survey}, where the a model is pre-trained on a upstreaming dataset and then the pre-trained model will be used as parameter initialization and continuing trained on datasets of various downstream tasks. The upstream task could be unsupervised task, such as masked language model \cite{devlin2018bert}\cite{liu2019roberta} for natural language processing, or supervised task, such as image classification for computer vision \cite{krizhevsky2012imagenet}. Pre-trained models could help the downstream task to achieve better training effect especially for the situation where the training data of the downstream task is limited. Meanwhile, compared with training dataset, the pre-trained model are usually more safely to be published. Therefore, thousands of pre-trained models has been published on model hub website such as HuggingFace \cite{wolf2019huggingface} , PyTorch-Hub \cite{jamieson2016non}, etc. On the other hand, previous work has demonstrated that for the same downstream task, the training performance may vary a lot when fine-tuned on different pre-trained models \cite{li2021palette}. Hence, selecting a good pre-trained model from model repository is an important step to achieve high training performance for the target task.

As the number of models in model repository becomes larger, it is compute infeasible to select a good model after fine-tuning all the models on the target task, and a few methods has been proposed to speed up the model selection process. We divide the proposed model selection methods into two categories, light-weight proxy score computation and model selection during fine-tuning. Specifically, for methods of light-weight proxy score computation, Task2Vec \cite{achille2019task2vec} embedding the upstream and downstream tasks into the same vector space, and models trained in upstream task closed to the downstream task could be selected directly; LEEP \cite{nguyen2020leep} is proposed to measure the transfer-ability for a pre-trained model on the target classification task; and in \cite{renggli2022model}, the KNN classifier is built on the hidden layer output of pre-trained models to approximate the training effect after fine-tuning. For methods of model selection during fine-tuning, Palette \cite{li2021palette} adopts successive halving \cite{jamieson2016non} to filter lower effect models at early training step, and Shift \cite{renggli2022shift} builds cost model to predict the training cost of successive halving and fine-tuning directly. The two-phase model selection framework could be viewed as combing the advantage of the two category methods. As the two-phase model selection framework is flexible, other model selection methods could be combined in this framework even in corresponding phase, such as we can also measure the model performance on benchmark datasets as \cite{zhai2019visual} and combine this strategy with LEEP \cite{nguyen2020leep} in coarse-recall phase, and we can also combine multi-model selection methods \cite{li2021palette}\cite{cui2021efficient}\cite{zhang2020efficient}\cite{ding2020boosting} in the fine-selection phase to achieve high ensemble performance. 

In this paper, we also propose to cluster models and mine convergence trend to speed up coarse-recall and fine-selection phase. The clustering and mining process both depend on the training performance of pre-trained models on benchmark datasets, such as GLUE \cite{wang2018glue} for natural language processing. As thousands of datasets are also published on the website, such as HuggingFace datasets \cite{wolf2019huggingface}, methods such as \cite{achille2019task2vec}\cite{feng2020language} could also be applied to measure the similarity between the task datasets and help to build more effective benchmark datasets which could cover a wider range of tasks for model selection.

\section{Future Work}
The future work could be summarized into three aspects. Firstly, the coarse-recall phase aims to retain the high performance pre-trained models in top models. A single proxy-score measurement may be not enough to return high performance models for different machine learning tasks. We plan to combine different light-weight tasks to return a high quality subset of models more robustly. Secondly, in this paper, we build benchmark datasets empirically. As a large of number of datasets has been published, we will study data-driven methods to build benchmark datasets which could cover more types of machine tasks, and meanwhile make benchmark datasets more compact to maintain performance matrix more cheaply. And thirdly, as model selection is an important step in the machine learning pipeline, the capability of automatically select high performance model enable us to build the whole machine learning pipeline for new task. We plan to build data management system which stores and maintains the pre-trained models and datasets, then support automatically selecting models efficiently to help users complete the model training for new task. 

\section{Conclusion}
In this paper, we propose a two-phase framework for fast model selection, where the coarse-recall phase implements light-weight proxy tasks to recall a much smaller number of candidate models and fine-selection phase only fine-tunes the models from this first phase to select the best model. To speed up these two phases, we build performance matrix by fine-tuning pre-trained models on benchmark datasets, and cluster models based on performance matrix to avoid duplicated proxy-score computation in coarse-recall phase and mine convergence trend to filter poorly-perform models more earlier in fine-selection phase. The experiments results on natural language process and computer vision tasks demonstrate the methods proposed in this paper could select a good model for the new task much faster compared with baseline methods.

\section{Acknowledgment}
This work was supported by the National Natural Science Foundation of China under Grant No. 62072458 and No. 62062058.
\bibliographystyle{IEEEtran}
\bibliography{output}

\clearpage
\appendix


\subsection{Mnli Results}
\begin{figure}[h] 
\centerline{\includegraphics[scale = 0.4]{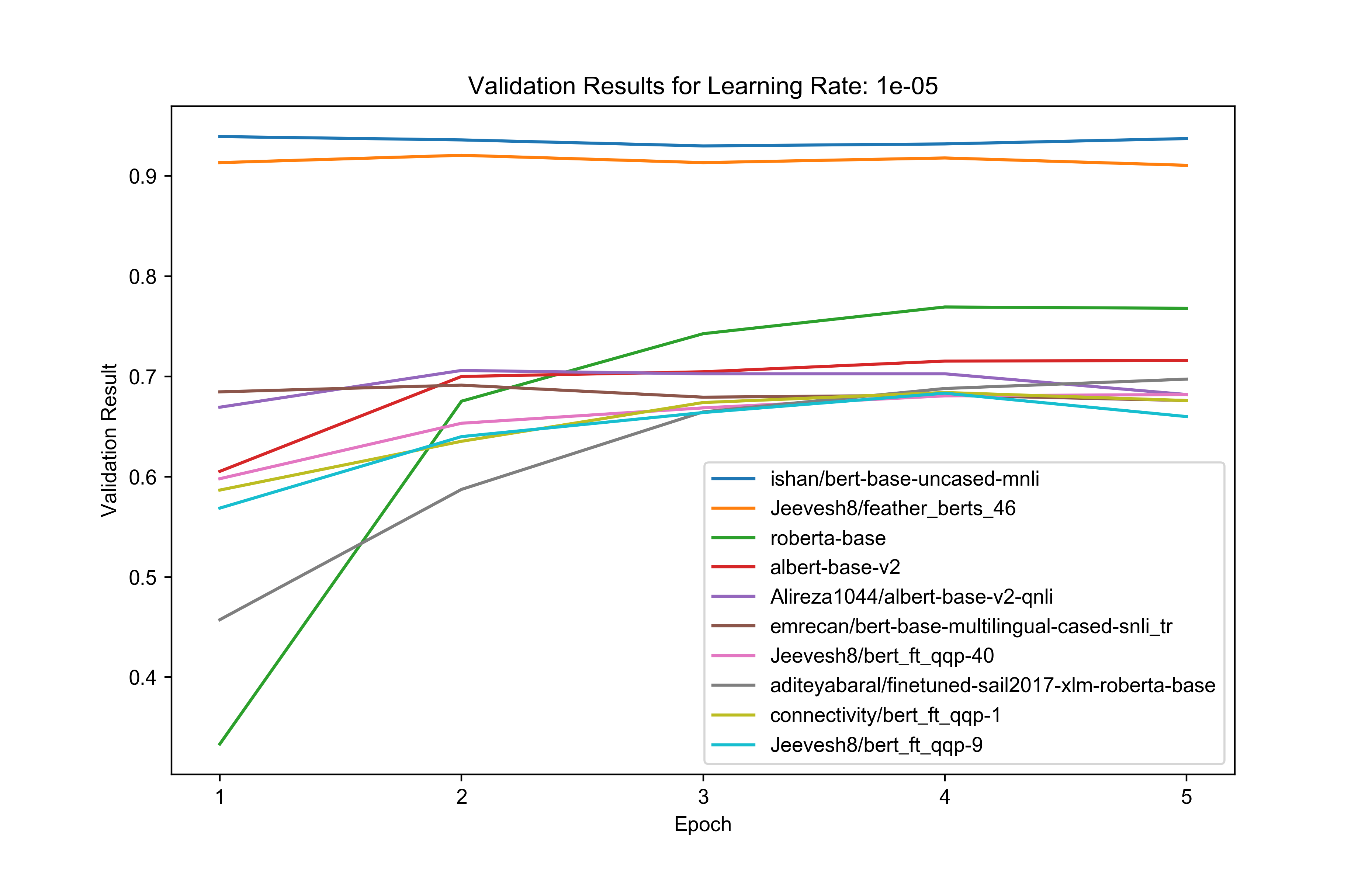}}
\caption{Top-10 models validation and test results on MNLI dataset. Learning rate is 1e-5, which is different than 3e-5 in the Fig. \ref{mnli_1}.} %
\label{mnli_2}
\end{figure}

We provide different models' validation results on MNLI dataset under a different learning rate 1e-5 in Fig. \ref{mnli_2}, compared to 3e-5 in Fig. \ref{mnli_1}. It can be observed that under new hyperparameter settings, the training dynamics of the models have changed. The performances of the top two models did not continuously decline with further training, suggesting a less severe overfitting issue. This indicates that the training process of models is highly sensitive to the setting of hyperparameters. In addition, we use our two-phase model selection method for the model training process under the new hyperparameters, and the performance and efficiency are consistent. Despite the changes in the training process, the variation in model performance was not significant enough to impact the effectiveness of our method. Therefore, our approach is robust to different hyperparameter settings in model training and is applicable across various model training scenarios.

\subsection{Model Details}
The pre-trained models we use are all from Huggingface's model hub \footnote{https://huggingface.co/models}. We list the full names of all the NLP and CV models used in our work in Table \ref{models_nlpcv}. Note that we sometimes use incomplete model names in the main text to save space by removing the name of the repository to which the model belongs. After removing the repository name prefix, the model names are still uniquely summarized in the list of models we use, so partial model names can also be used to pinpoint the corresponding model.

NLP models and CV models are listed in Table \ref{dataset} in total. All models are available using "https://huggingface.co/" as prefix.

\begin{table*}[htbp]
\begin{center}
\caption{NLP and CV Models}
\begin{tabular}{|c|c|}
\hline
\textbf{NLP model name} & \textbf{CV model name} \\
\hline
18811449050/bert\_finetuning\_test & facebook/deit-base-patch16-224 \\
\hline
aditeyabaral/finetuned-sail2017-xlm-roberta-base &  facebook/deit-base-patch16-384 \\
\hline
albert-base-v2 & facebook/deit-small-patch16-224 \\
\hline
aliosm/sha3bor-metre-detector-arabertv2-base & facebook/dino-vitb16 \\
\hline
Alireza1044/albert-base-v2-qnli & facebook/dino-vitb8 \\
\hline
anirudh21/bert-base-uncased-finetuned-qnli & facebook/dino-vits16 \\
\hline
aviator-neural/bert-base-uncased-sst2 & facebook/vit-msn-base \\
\hline
aychang/bert-base-cased-trec-coarse & facebook/vit-msn-small \\
\hline
bert-base-uncased & google/vit-base-patch16-224 \\
\hline
bondi/bert-semaphore-prediction-w4 & google/vit-base-patch16-384 \\
\hline
CAMeL-Lab/bert-base-arabic-camelbert-da-sentiment & google/vit-base-patch32-224-in21k \\
\hline
CAMeL-Lab--bert-base-arabic-camelbert-mix-did-nadi & lixiqi/beit-base-patch16-224-pt22k-ft22k-finetuned-FER2013-6e-05 \\
\hline
classla/bcms-bertic-parlasent-bcs-ter & lixiqi/beit-base-patch16-224-pt22k-ft22k-finetuned-FER2013-7e-05 \\
\hline
connectivity/bert\_ft\_qqp-1 & lixiqi/beit-base-patch16-224-pt22k-ft22k-finetuned-FER-5e-05-3 \\
\hline
connectivity/bert\_ft\_qqp-17 & microsoft/beit-base-patch16-224 \\
\hline
connectivity/bert\_ft\_qqp-7 & microsoft/beit-base-patch16-224-pt22k \\
\hline
connectivity/bert\_ft\_qqp-96 & microsoft/beit-base-patch16-224-pt22k-ft22k \\
\hline
dhimskyy/wiki-bert & microsoft/beit-base-patch16-384 \\
\hline
distilbert-base-uncased & microsoft/beit-large-patch16-224-pt22k \\
\hline
DoyyingFace/bert-asian-hate-tweets-asian-unclean-freeze-4 & mrgiraffe/vit-large-dataset-model-v3 \\
\hline
emrecan/bert-base-multilingual-cased-snli\_tr & sail/poolformer\_m36 \\
\hline
gchhablani/bert-base-cased-finetuned-rte & sail/poolformer\_m48 \\
\hline
gchhablani/bert-base-cased-finetuned-wnli & sail/poolformer\_s36 \\
\hline
ishan/bert-base-uncased-mnli & shi-labs/dinat-base-in1k-224 \\
\hline
jb2k/bert-base-multilingual-cased-language-detection & shi-labs/dinat-large-in22k-in1k-224 \\
\hline
Jeevesh8/512seq\_len\_6ep\_bert\_ft\_cola-91 & shi-labs/dinat-large-in22k-in1k-384 \\
\hline
Jeevesh8/6ep\_bert\_ft\_cola-47 & Visual-Attention-Network/van-base \\
\hline
Jeevesh8/bert\_ft\_cola-88 & Visual-Attention-Network/van-large \\
\hline
Jeevesh8/bert\_ft\_qqp-40 & oschamp/vit-artworkclassifier \\
\hline
Jeevesh8/bert\_ft\_qqp-68 & nateraw/vit-age-classifier \\
\hline
Jeevesh8/bert\_ft\_qqp-9 & - \\
\hline
Jeevesh8/feather\_berts\_46 & - \\
\hline
Jeevesh8/init\_bert\_ft\_qqp-24 & - \\
\hline
Jeevesh8/init\_bert\_ft\_qqp-33 & - \\
\hline
manueltonneau/bert-twitter-en-is-hired & - \\
\hline
roberta-base & - \\
\hline
socialmediaie/TRAC2020\_IBEN\_B\_bert-base-multilingual-uncased & - \\
\hline
Splend1dchan/bert-base-uncased-slue-goldtrascription-e3-lr1e-4 & - \\
\hline
XSY/albert-base-v2-imdb-calssification & - \\
\hline
Guscode/DKbert-hatespeech-detection & - \\
\hline

\end{tabular}
\label{models_nlpcv}
\end{center}
\end{table*}

\subsection{Dataset Details}NLP datasets and CV datasets are listed in Table \uppercase\expandafter{\romannumeral9}. Some datasets contains multiple subsets. All datasets are available using "https://huggingface.co/" as prefix. GLUE and SuperGLUE are the most common benchmark datasets in NLP. Cifar10 and MNIST are the most common benchmark datasets in CV. Other NLP datasets are described below:

\begin{itemize}
    \item \textbf{LysandreJik/glue-mnli-train} This datasets contain labelled MNLI dataset. The original MNLI dataset in GLUE does not have label, and the label is necessary for our experiment. This task is to predict the relation between the premise and the hypothesis. The result could be entailment, contradiction, or neutral. The labels of this dataset are balanced.
    \item \textbf{SetFit/qnli} This datasets contain labelled qnli dataset. The original qnli dataset in GLUE does not have label, and the label is necessary for our experiment. This task is to predict whether or not the paragraph contains the answer to the question. The labels of this dataset are balanced.
    \item \textbf{xnli} This dataset contains part of MNLI dataset after translated into different languages. The labels of this dataset are balanced.
    \item \textbf{stsb\_multi\_mt} This task is to score the similarity between two sentences on the scale of 0 to 5. The labels of this dataset are not balanced.
    \item \textbf{anli} This task is the same as MNLI dataset. However, the dataset is collected in an adversarial procedure. The labels of this dataset are not balanced.
    \item \textbf{tweet\_eval} This is a sentiment analysis task. The dataset is collected from Tweeter. The labels of this dataset are not balanced.
    \item \textbf{paws} This is a paraphrase identification task. The labels of this dataset are not balanced.
    \item \textbf{financial\_phrasebank} This is a sentiment analysis task in the realm of finance. The dataset is collected from financial news. The labels of this dataset are not balanced.
    \item \textbf{yahoo\_answers\_topics} This is a classification task. The dataset is collected from Yahoo. The labels of this dataset are balanced.
\end{itemize}

Other CV datasets are described below:

\begin{itemize}
    \item \textbf{food101} This dataset contains 101 kinds of food that need to predict. The size of the image is not the same. The labels of this dataset are balanced.
    \item \textbf{nelorth/oxford-flowers} This dataset contains 102 kinds of flowers that need to predict. The size of the images is not the same. The labels of this dataset are not balanced.
    \item \textbf{Matthijs/snacks} This dataset contains 20 kinds of snacks that need to predict. The size of the images is not the same. The labels of this dataset are slightly unbalanced.
    \item \textbf{beans} This dataset contains 3 kinds of leaves that need to predict. The size of the images is the same. The labels of this dataset are balanced.
    \item \textbf{cats\_vs\_dogs} This dataset contains images of cats or dogs and is a subset of Asirra dataset. The size of the images is not the same. The labels of this dataset are balanced.
    \item \textbf{trpakov/chest-xray-classification} This dataset contains images of chest x-ray. The size of the images is the same. The labels of this dataset are not balanced.
    \item \textbf{alkzar90/CC6204-Hackaton-Cub-Dataset} This daatset contains images of birds. The size of the images is not the same. The labels of this dataset are not balanced.
    \item \textbf{albertvillanova/medmnist-v2} This dataset contains images about biomedical. The size of the image is the same. The labels of this dataset are not balanced.
\end{itemize}

\begin{table}[t]
\centering
\caption{NLP and CV Datasets}
\begin{tabular}{cc}
\hline
\textbf{NLP dataset name} & \textbf{CV dataset name} \\
\hline
glue & food101 \\
super\_glue & nelorth/oxford-flowers \\
LysandreJik/glue-mnli-train & Matthijs/snacks \\
SetFit/qnli & beans \\
xnli & cats\_vs\_dogs \\
stsb\_multi\_mt & trpakov/chest-xray-classification \\
anli & cifar10 \\
tweet\_eval & MNIST \\
paws & alkzar90/CC6204-Hackaton-Cub-Dataset \\
financial\_phrasebank & albertvillanova/medmnist-v2 \\
yahoo\_answers\_topics & - \\
\hline
\end{tabular}
\label{dataset}
\end{table}

\subsection{Experiment on the Number of Dimensions for Max Average Error}
As discussed in Eq. \ref{sim} and Section V.B., we use top-k maximum average error to measure the model similarity and the parameter $k$ may influence the performance of the model selection algorithm. Thus, we test different values of $k$ while fixing other items. Due to the number of datasets, we choose $k=5,10,15$ for NLP clustering evaluation and $k=3,4,5$ for CV clustering evaluation. The result is shown in Table \ref{parameter k}. We can find that the influence of parameter $k$ is limited since the silhouette coefficient fluctuates within an acceptable range. Considering that the parameter $k$ in Eq. \ref{sim} should be able to filter noise and retain valid information, we choose $k=5$ in both tasks.

\begin{table}[]
\caption{Parameter K Selection}
\begin{tabular}{ccccccc}
\hline
                       & \multicolumn{3}{c}{NLP} & \multicolumn{3}{c}{CV} \\ \hline
K Value                & 5      & 10     & 15    & 3      & 4     & 5     \\
Silhouette Coefficient & 0.543  & 0.503  & 0.535 & 0.850  & 0.828 & 0.821 \\ \hline
\end{tabular}
\label{parameter k}
\end{table}

\subsection{Model cards}

\begin{figure*}[t] 
\centerline{\includegraphics[scale = 0.35]{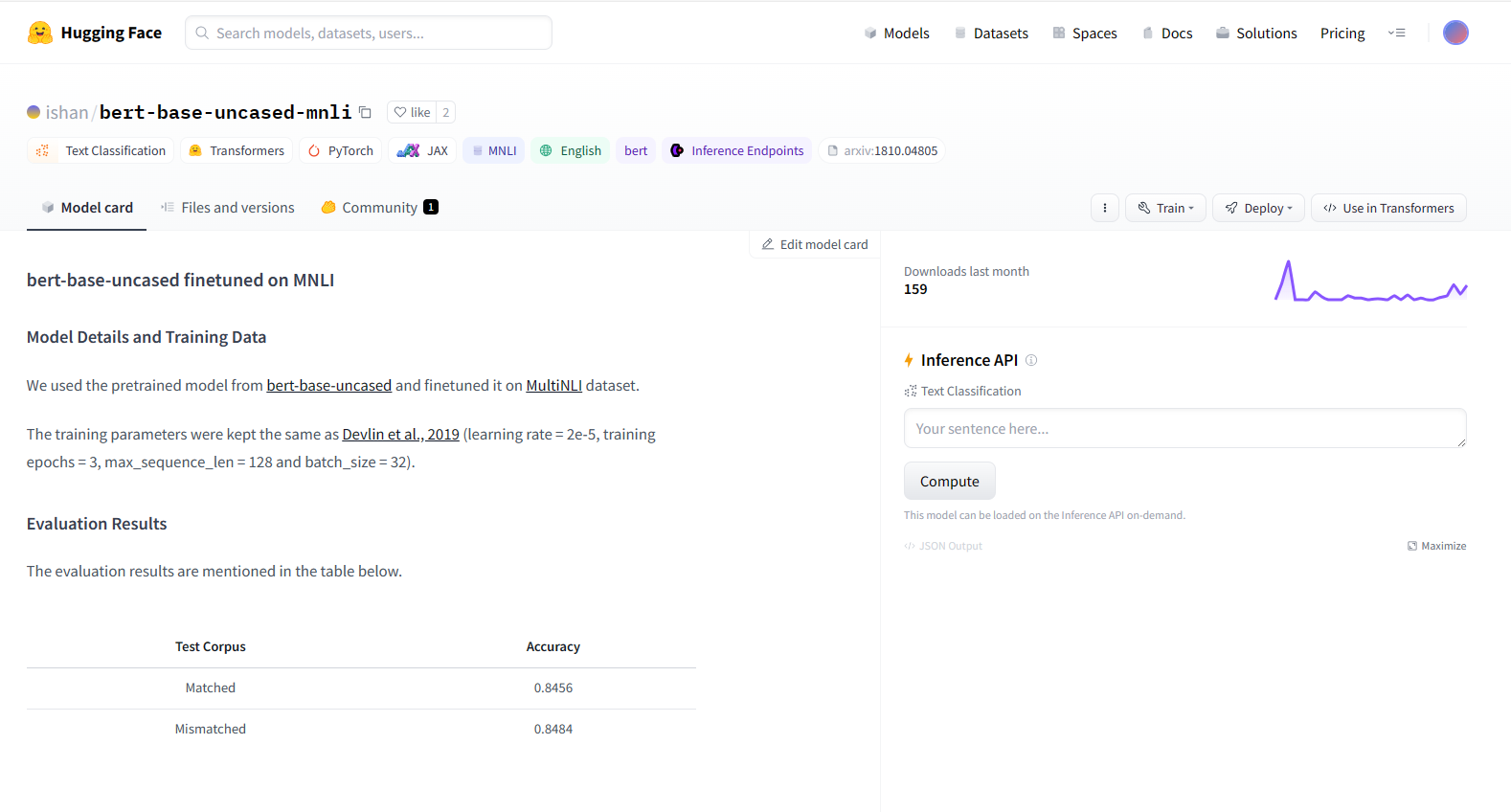}}
\caption{Model card of bert-base-uncased-mnli. Each model on HuggingFace has a model card to describe the model.} %
\label{mnli_model_card}
\end{figure*}

A model card is given in Fig. \ref{mnli_model_card}. A model card contain the general description of the model, such as structure and training information. 

\subsection{K-means Clustering Results}

The result of K-means clustering is shown in Table \ref{K-means clustering result}. This table is related to Table \ref{hierarchical clustering result} in section V. B. Model Clustering. In that section, we explain the result of hierarchical clustering in detail. We conclude that the result of hierarchical clustering is effective since the in-cluster models share the same model structure or training dataset while the silhouette coefficient is high. Here we give the result of K-means clustering to better prove our conclusion. Both the NLP clustering result and CV clustering result of the K-means clustering algorithm show less connection between in-cluster models. In the NLP part, the 2 biggest clusters, $C_2$ and $C_8$, consist of a mix of models that have different structures and training datasets. In the CV part, there is a cross mixing in $C_6$ and $C_7$, and the biggest cluster, $C_4$, does not show consistency in either model structure or training dataset. Thus, we take the method of hierarchical clustering as the main line of this paper.

\begin{table*}[htbp]
\begin{center}
\caption{Model Clustering Results Using K-means}
\begin{tabular}{|c|c|c|}
\hline
\multicolumn{3}{|c|}{\textbf{Model Clusters of Natural Language Processing}} \\
\hline
\textbf{Cluster} & \textbf{Size}& \textbf{Pre-trained Models} \\
\hline
$C_1$ & 2 & \makecell{gchhablani--bert-base-cased-finetuned-rte, anirudh21--bert-base-uncased-finetuned-qnli}  \\
\hline
$C_2$ & 5 & \makecell{Jeevesh8--bert\_ft\_cola-88, DoyyingFace--bert-asian-hate-tweets-asian-unclean-freeze-4, \\ bert-base-uncased , aditeyabaral--finetuned-sail2017-xlm-roberta-base, Jeevesh8--512seq\_len\_6ep\_bert\_ft\_cola-91} \\
\hline
$C_3$ & 2 & \makecell{manueltonneau--bert-twitter-en-is-hired, aychang--bert-base-cased-trec-coarse} \\
\hline
$C_4$ & 2 & \makecell{XSY--albert-base-v2-imdb-calssification, distilbert-base-uncased
} \\
\hline
$C_4$ & 4 & \makecell{ishan--bert-base-uncased-mnli, Alireza1044--albert-base-v2-qnli, \\ albert-base-v2, Jeevesh8--feather\_berts\_46 : 
} \\
\hline
$C_5$ & 2 & \makecell{CAMeL-Lab--bert-base-arabic-camelbert-mix-did-nadi, aliosm--sha3bor-metre-detector-arabertv2-base} \\
\hline
$C_6$ & 3 & \makecell{socialmediaie--TRAC2020\_IBEN\_B\_bert-base-multilingual-uncased, jb2k--bert-base-multilingual-cased-language-detection, \\ emrecan--bert-base-multilingual-cased-snli\_tr}  \\
\hline
$C_7$ & 2 & \makecell{dhimskyy--wiki-bert, bondi--bert-semaphore-prediction-w4} \\
\hline
$C_8$ & 5 & \makecell{Jeevesh8--init\_bert\_ft\_qqp-33, Jeevesh8--bert\_ft\_qqp-68, Jeevesh8--bert\_ft\_qqp-40, \\ connectivity--bert\_ft\_qqp-1, Jeevesh8--bert\_ft\_qqp-9} \\
\hline
$C_9$ & 4 & \makecell{connectivity--bert\_ft\_qqp-96, connectivity--bert\_ft\_qqp-7, \\ connectivity--bert\_ft\_qqp-17, Jeevesh8--init\_bert\_ft\_qqp-24} \\
\hline
$C_{10}$ & 2 & \makecell{Splend1dchan--bert-base-uncased-slue-goldtrascription-e3-lr1e-4, Jeevesh8--6ep\_bert\_ft\_cola-47} \\
\hline
\multicolumn{3}{|c|}{\textbf{Model Clusters of Computer Vision}} \\
\hline
\textbf{Cluster} & \textbf{Size}& \textbf{Pre-trained Models} \\
\hline
$C_1$ & 6 & \makecell{shi-labs/dinat-large-in22k-in1k-224, shi-labs/dinat-large-in22k-in1k-384, microsoft/beit-base-patch16-224-pt22k-ft22k, \\ lixiqi/beit-base-patch16-224-pt22k-ft22k-finetuned-FER2013-7e-05, \\ lixiqi/beit-base-patch16-224-pt22k-ft22k-finetuned-FER2013-6e-05, \\ lixiqi/beit-base-patch16-224-pt22k-ft22k-finetuned-FER-5e-05-3} \\
\hline
$C_2$ & 2 & \makecell{nateraw/vit-age-classifier, facebook/dino-vitb16} \\
\hline
$C_3$ & 3 & \makecell{sail/poolformer\_m48, sail/poolformer\_m36, sail/poolformer\_s36} \\
\hline
$C_4$ & 7 & \makecell{facebook/vit-msn-small, facebook/vit-msn-base, facebook/deit-base-patch16-384, \\ google/vit-base-patch32-224-in21k, Visual-Attention-Network/van-large, facebook/deit-base-patch16-224, facebook/dino-vits16} \\
\hline
$C_5$ & 4 & \makecell{Visual-Attention-Network/van-base, microsoft/beit-large-patch16-224-pt22k, \\ facebook/deit-small-patch16-224, shi-labs/dinat-base-in1k-224} \\
\hline
$C_6$ & 2 & \makecell{microsoft/beit-base-patch16-384, google/vit-base-patch16-384} \\ 
\hline
$C_7$ & 2 & \makecell{microsoft/beit-base-patch16-224, google/vit-base-patch16-224} \\ 
\hline
\end{tabular}
\label{K-means clustering result}
\end{center}
\end{table*}

\end{document}